\documentclass[runningheads]{llncs}

\usepackage{eccv}

\usepackage{eccvabbrv}

\usepackage[dvipsnames]{}
\usepackage{xcolor,colortbl}

\usepackage{times}
\usepackage{epsfig}
\usepackage{graphicx}
\usepackage{amsmath}
\usepackage{amssymb}
\usepackage{dsfont}
\usepackage{tabularx}
\usepackage{bibunits}

\usepackage{graphicx}
\usepackage{booktabs}

\usepackage{enumitem}
\usepackage{amsmath}
\usepackage{adjustbox}
\usepackage{listings}
\usepackage{times}
\usepackage{epsfig}
\usepackage{graphicx}
\usepackage{amsmath}
\usepackage{amssymb}
\usepackage{wrapfig}
\usepackage[font=footnotesize]{caption}
\usepackage{subcaption}
\usepackage{float}
\usepackage{array}
\usepackage{color}
\usepackage{multirow}
\usepackage{diagbox}
\usepackage{booktabs}
\usepackage{siunitx}
\usepackage{color}
\usepackage{graphbox}
\usepackage[table]{}
\usepackage{comment}
\usepackage{wrapfig}
\usepackage[normalem]{ulem}
\usepackage{tikz}
\usepackage{multirow}
\usepackage{varwidth}
\usepackage{pifont} %
\usepackage{enumitem}
\definecolor{pink}{RGB}{213,126,190}
\definecolor{orange}{RGB}{239,133,54}
\definecolor{lightblue}{RGB}{173,216,230}
\definecolor{darkblue}{RGB}{8,81,156}
\definecolor{stanfordgrey}{RGB}{46,45,41}
\definecolor{cardinalred}{RGB}{253,141,60}
\definecolor{codegreen}{rgb}{0,0.4,0}
\definecolor{codegray}{rgb}{1.0,0.5,0.5}
\definecolor{codepurple}{rgb}{0.58,0,0}
\definecolor{tealblue}{rgb}{0,0.5,0.5}
\definecolor{codebackcolour}{rgb}{0.95,0.95,0.92}
\definecolor{darkgreen}{RGB}{0,127,0}
\definecolor{darkred}{RGB}{200,0,0}
\definecolor{orange}{rgb}{1,0.5,0}
\definecolor{brickred}{rgb}{0.8, 0.25, 0.33}
\definecolor{bananayellow}{rgb}{1.0, 0.88, 0.21}
\definecolor{falured}{rgb}{0.5, 0.09, 0.09}
\definecolor{teal}{rgb}{0.1176, 0.5019, 0.5019}

\definecolor{tabfirst}{rgb}{1, 1, 1} % orange
\definecolor{tabsecond}{rgb}{1, 1, 1} % yellow

\definecolor{tabthird}{rgb}{1,1,1} % red
\newcommand{\xmark}{\ding{55}}%
\newcolumntype{C}{>{\centering\arraybackslash}X}
\newlength\savewidth

\usepackage[accsupp]{axessibility}  % Improves PDF readability for those with disabilities.

\usepackage[pagebackref,breaklinks,colorlinks,citecolor=eccvblue]{hyperref}
\usepackage{orcidlink}

\begin{document}

\title{NeRF-MAE: Masked AutoEncoders for Self-Supervised 3D Representation Learning for Neural Radiance Fields} 

\titlerunning{NeRF-MAE}

\author{Muhammad Zubair Irshad\inst{1}, \inst{2}\orcidlink{0000-0002-1955-6194} \and
Sergey Zakharov\inst{1}\orcidlink{0000-0002-6231-6137} \and
Vitor Guizilini\inst{1}\orcidlink{0000-0002-8715-8307} \and
\\ Adrien Gaidon\inst{1}\orcidlink{0000-0001-8820-550X} \and
Zsolt Kira\inst{2}\orcidlink{0000-0002-2626-2004} \and
Rares Ambrus\inst{1}\orcidlink{0000-0002-3111-3812}
\vspace{0.20cm}
\\$^1$
\hspace{.1cm}Toyota Research Institute \ \ $^2$\hspace{.1cm}Georgia Tech\\
}

\authorrunning{M.Z.~Irshad et al.}

\institute{}
% \institute{Toyota Research Institute \email{\{first.last\}@tri.global} \and Georgia Tech \email{\{zkira\}@gatech.edu}}

\maketitle

\begin{center}\vspace{-0.5cm}
\href{https://nerf-mae.github.io/}{nerf-mae.github.io}
\vspace{-0.08cm}

\end{center}

\begin{figure}
   \vspace{-0.7cm}
   \centering
       \includegraphics[width=1.0\textwidth]{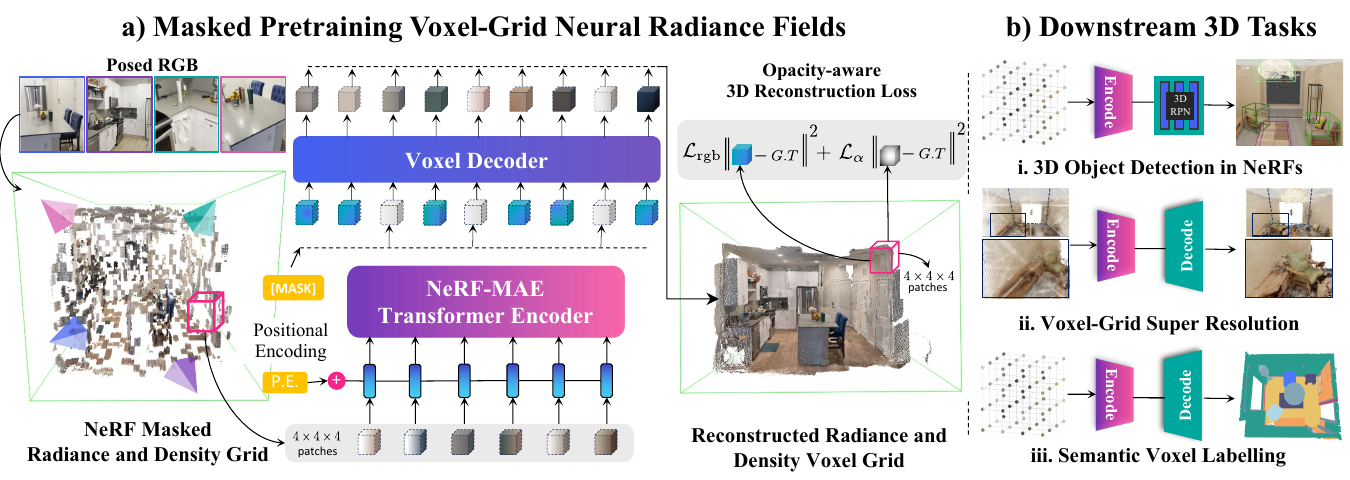}
    \caption{\textbf{NeRF-MAE Overview:} The first large-scale self-supervised pretraining utilizing NeRF's radiance and density grid as an input modality. Our approach uses standard Transformers to learn powerful 3D representations in~(\textbf{a}) an opacity-aware dense volumetric masked self-supervised learning objective.~(\textbf{b}) when fine-tuned on a small subset of data, our representation improves many 3D downstream tasks such as 3D object detection, super-resolution, and voxel-labeling.
      }
   \label{fig:teaser}
   \vspace{-0.5cm}
\end{figure}

\vspace{-0.2cm}

\begin{abstract}
\vspace{-16pt}
Neural fields excel in computer vision and robotics due to their ability to understand the 3D visual world such as inferring semantics, geometry, and dynamics. Given the capabilities of neural fields in densely representing a 3D scene from 2D images, we ask the question: Can we scale their self-supervised pretraining, specifically using masked autoencoders, to generate effective 3D representations from posed RGB images. Owing to the astounding success of extending transformers to novel data modalities, we employ standard 3D Vision Transformers to suit the unique formulation of NeRFs. We leverage NeRF's volumetric grid as a dense input to the transformer, contrasting it with other 3D representations such as pointclouds where the information density can be uneven, and the representation is irregular. Due to the difficulty of applying masked autoencoders to an implicit representation, such as NeRF, we opt for extracting an explicit representation that canonicalizes scenes across domains by employing the camera trajectory for sampling. Our goal is made possible by masking random patches from NeRF's radiance and density grid and employing a standard 3D Swin Transformer to reconstruct the masked patches. In doing so, the model can learn the semantic and spatial structure of complete scenes. We pretrain this representation at scale on our proposed curated posed-RGB data, totaling over 1.8 million images. Once pretrained, the encoder is used for effective 3D transfer learning. Our novel self-supervised pretraining for NeRFs, NeRF-MAE, scales remarkably well and improves performance on various challenging 3D tasks. Utilizing unlabeled posed 2D data for pretraining, NeRF-MAE significantly outperforms self-supervised 3D pretraining and NeRF scene understanding baselines on Front3D and ScanNet datasets with an absolute performance improvement of over 20\% AP50 and 8\% AP25 for 3D object detection.

\end{abstract}    
\section{Introduction}
\label{sec:intro}

\looseness=-1
Neural Radiance Fields~(NeRFs~\cite{mildenhall2020nerf, barron2021mip}) have exhibited impressive capability in synthesizing novel views by utilizing posed RGB images as input. Their ability to capture scenes~\cite{wang2021neus, sucar2021imap, ost2021neural} with high fidelity has led to a surge in NeRF-related works, using them for AR/VR~\cite{kuang2022neroic, nguyen2024semantically}, robotics~\cite{irshad2022shapo, yen2022nerf}, and driving~\cite{tancik2022block, irshad2023neo360} applications. 

Most importantly, neural fields' emergent properties~\cite{xie2021neural} have made them suitable for tasks beyond showcasing higher rendering quality. Prior approaches have successfully applied neural fields to extract accurate scene geometry~\cite{wang2021neus, yariv2023bakedsdf, OechsleICCV2019}, estimate camera poses~\cite{yenchen2021inerf, wang2021nerfmm}, infer accurate semantics~\cite{cen2023segment, gaussian_grouping}, and learn 3D correspondences~\cite{yen2022nerfsupervision, simeonovdu2021ndf}. NeRFs have also emerged as a promising 3D data storage medium~\cite{jeong2022perfception}, and have proven to outperform traditional supervised approaches~\cite{jeong2022perfception, hu2023nerf} for challenging 3D tasks like object detection~\cite{hu2023nerf, xu2023nerf} and instance segmentation~\cite{liu2023instance}.

\looseness=-1
Furthermore, Vision Transformers~(ViT)~\cite{Dosovitskiy17} have also transformed computer vision in recent years~\cite{he2021masked, xie2022simmim, baevski2022data2vec}. Their unique approach, encoding visual information from patches and leveraging self-attention blocks for modeling long-range global information~\cite{raghu2021vision, zhai2022scaling, mao2021voxel}, sets them apart from conventional CNNs. ViTs, notably when combined with recent advances in pretraining strategies like masked auto-encoders (MAE)~\cite{he2021masked, zhang2023learning, yu2022point}, outperform CNNs in robust feature representation learning, demonstrating state-of-the-art results when fine-tuned for downstream tasks~\cite{raghu2021vision, tang2022self, xie2021self}.

\begin{figure}[b]
    \centering
    \resizebox{1.0\linewidth}{!}{%
\includegraphics{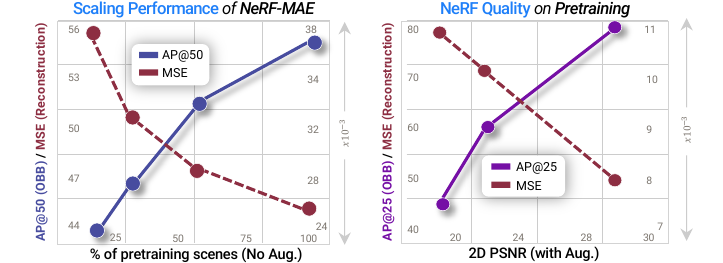}
    }
\caption{\textbf{Quantitative results} showing our representation improves with more unsupervised data and as well as better reconstruction quality of NeRFs.}
    \label{fig:quantitative_results_scaling_laws}
\end{figure}

Because of NeRF's inherent ability to densely represent a 3D scene from 2D images, this paper explores scaling up their self-supervised pretraining~(as we show in Fig.~\ref{fig:quantitative_results_scaling_laws} and Sec.~\ref{sec:experiments}) through the use of masked autoencoders. The notion of masked auto-encoding is equally suited to NeRF's radiance and density grid as to images~(Fig.~\ref{fig:comparison_mae}). As opposed to other 3D representations such as meshes, point clouds, or lidar data which only model surface-level information, are highly irregular data structures and the information density can be extremely uneven~\cite{pang2022masked}, NeRF's radiance and density grid is similar in principle to 2D images for masked auto-encoding. This is the case because it provides high information density, spatial data redundancy~\cite{he2021masked}, and a regular grid ensuring unbiased sampling between occupied and unoccupied areas. Owing to these nice properties, the radiance and density grid obtained from NeRFs have shown to achieve state-of-the-art scene understanding such as segmentation and detection~\cite{hu2023nerf, liu2023instance, xu2023nerf}.

Driven by this analysis, we propose a new framework termed~\textbf{NeRF-MAE}~(\textbf{NeRF} \textbf{M}asked~\textbf{A}uto~\textbf{E}ncoders) for self-supervised learning directly within Neural Radiance Fields. Utilizing NeRF's radiance and density grid as an input modality, NeRF-MAE employs standard 3D Transformers in a masked autoencoding objective suited to the unique formulation of NeRFs. In comparison to prior generalizable NeRF or 3D representation learning methods~(Table~\ref{tab:prior_approaches_comparison}), NeRF-MAE offers the following distinct advantages:~\textbf{1)} The masked self-supervised learning objective is directly enforced on dense regular grids with high-density information offering data redundancy, thus enabling the use of standard Transformer architecture without requiring careful adaptation or adopting extra non-Transformers Blocks~(such as DGCNN~\cite{dgcnn} used in Point-BERT~\cite{yu2022point}.~\textbf{2)} By disentangling representation learning and NeRF training, our model can effectively utilize a large amount of data from diverse sources. This is in contrast to most previous generalizable NeRF methods that only train on a single type of data source for priors.

\captionsetup[table]{skip=6pt}
\begin{table}[t]
\normalsize
\centering
\resizebox{\textwidth}{!}{
\begin{tabular}{c|cccccc|c}
\toprule
% ~\textbf{Method} & PixelNeRF~\cite{yu2020pixelnerf} & NeRFRPN~\cite{hu2023nerf} & PerFception~\cite{jeong2022perfception}& Voxel-MAE~\cite{hess2022masked}& PiMAE~\cite{chen2023pimae} & SUNetR~\cite{tang2022self}&\cellcolor{yellow!40}~\textbf{NeRF-MAE}  \\
% \midrule
% \# Scenes & 40 & 515 & 1500 & 850 & -&5000&\cellcolor{yellow!40}3637 \\
% \# Frames & 4k & 80k & 2.5M & - & 10k& - & \cellcolor{yellow!40}1.65M \\
% S.S. Pretraining & \textcolor{falured}{\xmark} & \textcolor{falured}{\xmark} & \textcolor{falured}{\xmark} & \textcolor{teal}{\checkmark} & \textcolor{teal}{\checkmark} & \textcolor{teal}{\checkmark} &\cellcolor{yellow!40}\textcolor{teal}{\checkmark} \\
% Input Data & RGB* & RGB* & RGB* & Pointcloud & RGB-D& CT Scan & \cellcolor{yellow!40}RGB* \\
% 3D Representation & NeRF & NeRF & NeRF & Voxel & Pointclouds&Voxel &\cellcolor{yellow!40}NeRF \\
% Data domain(s) & Single~(\textcolor{falured}{1}) & Single~(1) & Single~(\textcolor{falured}{1}) & Single~(\textcolor{falured}{1}) & Single~(\textcolor{falured}{1}) & Multiple~(\textcolor{teal}{5}) &\cellcolor{yellow!40}Multiple~(\textcolor{teal}{4})  \\

~\textbf{Method} & PixelNeRF~\cite{yu2020pixelnerf} & NeRFRPN~\cite{hu2023nerf} & PerFception~\cite{jeong2022perfception}& Voxel-MAE~\cite{hess2022masked}& PiMAE~\cite{chen2023pimae} & SUNetR~\cite{tang2022self}&\textbf{NeRF-MAE}  \\
\midrule
\# Scenes & 40 & 515 & 1500 & 850 & -&5000&3668 \\
\# Frames & 4k & 80k & 2.5M & - & 10k& - & 1.8M \\
S.S. Pretraining & \textcolor{falured}{\xmark} & \textcolor{falured}{\xmark} & \textcolor{falured}{\xmark} & \textcolor{teal}{\checkmark} & \textcolor{teal}{\checkmark} & \textcolor{teal}{\checkmark} &\textcolor{teal}{\checkmark} \\
Input Data & RGB* & RGB* & RGB* & Pointcloud & RGB-D& CT Scan & RGB* \\
3D Representation & NeRF & NeRF & NeRF & Voxel & Pointclouds&Voxel &NeRF \\
Data domain(s) & Single~(\textcolor{falured}{1}) & Single~(1) & Single~(\textcolor{falured}{1}) & Single~(\textcolor{falured}{1}) & Single~(\textcolor{falured}{1}) & Multiple~(\textcolor{teal}{5}) &Multiple~(\textcolor{teal}{4})  \\

\bottomrule
\end{tabular}
}
\caption{Specifications of supervised or self-supervised approaches to 3D representation or generalizable learning, highlighting our novel self-supervised~(S.S) 3D pretraining strategy for NeRFs which requires posed RGB images as input. * denotes multi-view posed RGB images. }
% Data domain refers to the combination of multiple data domains on which the published model has been trained on.}

\label{tab:prior_approaches_comparison}
% \vspace{-10mm}
\end{table}

% \zk{Will we compare to these methods in the tables?}.
% \zk{Why is number of frames dashed line for SUNetR/Voxel-MAE? Can we not get that information? }
\looseness=-1
As shown in Fig.~\ref{fig:teaser}, NeRF-MAE consists of a volumetric masking and embedding module and an auto-encoder. Due to the challenges of applying masked autoencoders to implicit representations like NeRF, we derive a grid of radiance and density that canonicalizes scenes across domains~(\S~\ref{subsec:nerf_4d_grid}) and utilizes camera trajectories for sampling. Subsequently, we partition the 4D voxel grid into smaller patches, randomly masking a substantial ratio to minimize data redundancy. The autoencoder's reconstruction objective is enforced to account for both density and radiance~(\S~\ref{subsec:masked_pretraining_nerf}), which learns high-level features from the unmasked volumetric patches and reconstructs the masked radiance and density patches in the canonical frame. Notably, our encoder's backbone consists of a standard 3D SwinTransformer, while a voxel decoder is employed for reconstruction.

\begin{figure}[t]
    \centering
    \resizebox{0.98\linewidth}{!}{%
\includegraphics{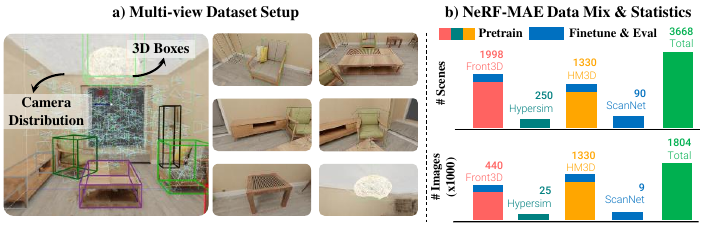}
    }
    \caption{\textbf{NeRF-MAE dataset mix:}~\textbf{a)} Multi-view dataset with camera distribution,~\textbf{b)} diverse scenes from different sources i.e. Front3D~\cite{fu20213d}, Hypersim~\cite{roberts2021hypersim}, HM3D~\cite{ramakrishnan2021hm3d} and ScanNet~\cite{dai2017scannet} totaling over 3600 scenes and 1.8M images used for pertaining and evaluating NeRF-MAE.
    }
    \label{fig:dataset_fig1}
\end{figure}

\looseness=-1
For training and evaluating our model, we curate a large-scale dataset for NeRF pretraining, encompassing 4 diverse sources in synthetic and real domains. Our dataset comprises multi-view posed images, corresponding NeRFs, and sampled radiance and density grids from Front3D, ScanNet, ARKitScenes, and Hypersim, accumulating over 1.8M images and 3600+ scenes~(Table.~\ref{tab:prior_approaches_comparison} and Figure~\ref{fig:dataset_fig1}). 
Our method proves effective, with pretrained models demonstrating strong generalization across a range of downstream tasks. Our approach, NeRF-MAE, achieves significant improvement over other self-supervised 3D pretraining baselines~(\S~\ref{sec:pretraining_3D_baelines}) as well as scene-understanding baselines~(\S~\ref{sec:comparsons_downstream_tasks}), specifically achieving a 21.5\% AP50 and 8\% AP25 improvement on Front3D and ScanNet 3D Object Bounding Box~(OBB) prediction while requiring less than half the data requirement of the state-of-the-art baseline to achieve the same performance. To summarize, we make the following contributions:
\begin{itemize}[noitemsep]
   \item To our knowledge, we introduce the first~\textbf{fully self-supervised transformer-based 3D pretraining} utilizing Neural Radiance Field's radiance and density grid as an input modality coupled with an~\textbf{opacity-aware masked reconstruction objective}.
   \item A~\textbf{large scale pretraining of Neural Radiance Fields} with robust evaluation on diverse datasets, over 1.8M images, and 3600+ indoor scenes, using~\textbf{a single model}.
   \item Our proposed approach~\textbf{significantly outperforms self-supervised 3D pretraining as well as NeRF scene understanding baselines} on multiple downstream 3D tasks, showing over 20\% AP50 and 12\% mAcc on 3D object detection and semantic voxel labeling respectively on Front3D dataset while~\textbf{requiring less than half the data} requirement of the state-of-the-art baseline to achieve the same performance.
\end{itemize}

\section{Related Works}
\label{sec:related_works}

\looseness=-1
\textbf{3D Representation Learning:} Supervised learning for 3D data~\cite{qi2017pointnet++, irshad2022shapo, yan2022shapeformer, irshad2022centersnap, he2022voxset} has made promising progress. Although powerful, these techniques require expensive 3D labels. Alternatively, self-supervised learning~\cite{yan2022implicit, liu2022masked, jiang2023self, zhang_depth_contrast, weinzaepfel2022croco} and masked autoencoders~\cite{he2021masked} learn strong representations for various tasks without needing labeled data. Follow-up works extended this idea to the 3D domain by reconstructing masked 3D coordinates with point clouds~\cite{yu2022point, chen2023pimae, pang2022masked, zhang2022point}, meshes~\cite{liang2022meshmae} or voxels~\cite{hess2022masked} as 3D representations. While successful, they require careful adaptation of existing architectures due to irregularity in data structures and uneven information density. Contrastingly, NeRF-MAE uses standard Transformers, while requiring readily available and cheaper-to-obtain posed 2D data compared to lidar scans or accurate mesh reconstructions.

\looseness=-1

\noindent \textbf{NeRF as 3D Scene Representation:} Advances in differentiable rendering~\cite{niemeyer2020differentiable, Chen2019, remelli2020meshsdf, zhang2021holistic, takikawa2021nglod, wang2021neus} have allowed for learning 3D representations using only image supervision. Neural Radiance Fields (NeRFs)~\cite{mildenhall2020nerf} is one example that excels in novel-view synthesis. Extensions of NeRF study generalizable scene modelling~\cite{irshad2023neo360, yu2020pixelnerf, chen2021mvsnerf, trevithick2021grf, wu2023reconfusion}, semantic and open-world 3D understanding~\cite{kundu2022panoptic, Siddiqui_2023_CVPR, Zhi:etal:ICCV2021, lerf2023, Engelmann2023openreno}, generative modelling~\cite{Niemeyer2020GIRAFFE, wang2022clipnerf, instructnerf2023}, robotics~\cite{zhu2022nice, sucar2021imap, yenchen2021inerf, Ze2023GNFactor, shen2023F3RM}, dynamic scenes~\cite{pumarola2020d, liu2020neural, yuan2021star, ost2020neural} and compositional representations~\cite{ost2020neural, yang2021objectnerf}. Recently, NeRFs have shown competitive performance for supervised 3D tasks such detection~\cite{hu2023nerf, xu2023nerf} and segmentation~\cite{liu2023instance} when compared with methods that require depth as input~\cite{rukhovich2022fcaf3d}. However, little attention has been given to representation learning for NeRFs. Our approach fills this gap by focusing on self-supervised pretraining for neural radiance fields, utilizing NeRF's dense radiance and density grid for pretraining to improve performance on various downstream 3D tasks.

\looseness=-1
\noindent \textbf{Vision Transformers:} Computer vision has undergone a transformative shift with the emergence of vision transformers~\cite{dosovitskiy2020image, raghu2021vision}. Their ability to model global and local contexts has sparked interest in exploring methods to pretrain these backbones. MoCoV3~\cite{chen2021mocov3} and MAE~\cite{he2021masked} delved into different aspects of self-supervised Vision Transformers. In particular, MAE~\cite{he2021masked} achieved state-of-the-art results by drawing inspiration from BERT~\cite{devlin2018bert}, which involves randomly masking words in sentences, and using masked image reconstruction. Subsequent research extended this concept to the 3D domain~\cite{pang2022masked, tang2022self, liang2022meshmae, hess2022masked}, leading to impressive results in downstream 3D vision tasks.

\section{NeRF-MAE: Self-Supervised Pretraining for NeRFs}

NeRF-MAE introduces a self-supervised framework aimed at enhancing 3D representation learning in the context of Neural Radiance Fields~\cite{mildenhall2020nerf}. Our approach allows learning robust 3D representations from posed RGB images, leading to substantial improvements in various downstream 3D applications, as we show in Sec.~\ref{sec:experiments}. As shown in Figures~\ref{fig:teaser} and~\ref{fig:dataset_fig2}, such a goal is made possible by two major components:~\textbf{(i)} An explicit 4D radiance and density grid extraction module in the canonical world frame using a camera-trajectory aware sampling from a fully-trained implicit NeRF model~(described in Sec.~\ref{subsec:nerf_4d_grid}) and~\textbf{(ii)} A masked self-supervised pretraining module, operating directly on the explicit NeRF's 4D radiance and density grid, to train a standard 3D SwinTransformer~\cite{liu2021swin, yang2023swin3d} encoder and a voxel decoder using an opacity-aware masked reconstruction objective in 3D~(discussed in Sec.~\ref{subsec:masked_pretraining_nerf}). Our NeRF pretraining dataset is described in Sec.~\ref{subsec:nerf_pt_dataset} and key preliminaries to understand our pipeline are detailed in Sec.~\ref{subsec:preliminaries}.

\begin{figure}[t]
    \centering
    \resizebox{1.0\linewidth}{!}{%
\includegraphics{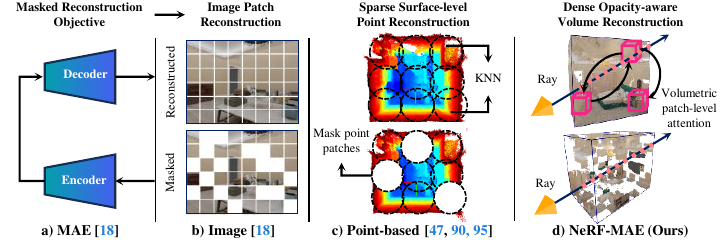}
    }
    \caption{\textbf{NeRF-MAE Comparison:} In       \S~\ref{sec:pretraining_3D_baelines}, we compare our method to point-based pretraining approaches, which are limited by their modeling of surface-level sparse points. In contrast, NeRF-MAE resembles images in terms of high information density and structural regularity, hence positioning it as a direct extension of image MAE to 3D. Leveraging NeRF's dense volumetric information and opacity-aware reconstruction loss, we achieve superior representation learning.
    }
    \label{fig:comparison_mae}
\end{figure}%

\subsection{Preliminaries - NeRF and InstantNGP:} Given posed 2D images, NeRF represents 
\label{subsec:preliminaries}
a 3D scene implicitly. It uses a neural network, $f(\mathbf{x}, \theta)$, to predict color ($\mathbf{c}_i$) and density ($\sigma_i$) at any given 3D query position~($x_i$) and viewing direction~($\theta_i$) as input. The 4D color and density outputs are used in an alpha compositing process to generate rendered images through volume rendering with near and far bounds $t_{n}$ and $t_{f}$, as highlighted in the equations below:
\begin{equation}
C(\mathbf{r})=\int_{t_n}^{t_f} T(t) \sigma(\mathbf{r}(t)) \mathbf{c}(\mathbf{r}(t), \mathbf{d}) d t
\end{equation}

\looseness=-1
\noindent where $T(t)=\exp \left(-\int_{t_n}^t \sigma(\mathbf{r}(s)) d s\right)$ and $\boldsymbol{r}$ denotes the camera ray. Although our work is suited to any input NeRF formulation, we choose Instant-NGP~\cite{muller2022instant} to model the radiance and density of a 3D scene using a small Multi-layer perceptron~(MLP) with a sparse occupancy grid. A fast ray marching routine is utilized by ray compacting based on occupancy grid values. We also utilize a multi-resolution hash encoding from~\cite{muller2022instant} for faster NeRF training. An example trained scene is shown in Figure~\ref{fig:dataset_fig2}~(b).

\subsection{Neural Radiance Field Grid Representation:}
\label{subsec:nerf_4d_grid}
\begin{figure}[t]
    \centering
    \resizebox{0.95\linewidth}{!}{%
\includegraphics{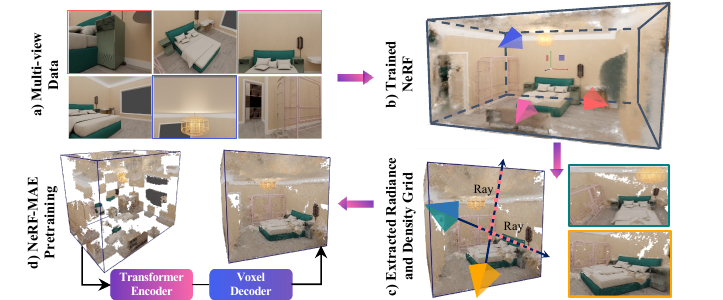}
    }
    \caption{\textbf{NeRF-MAE data processing flow for pretraining:} showing~\textbf{a)} multi-view training data,~\textbf{b)} trained NeRF representation,~\textbf{c)} extracted radiance and density grid from NeRF and~\textbf{d)} masked pretraining of the radiance and density voxel-grid neural radiance field.
    }
    
    \label{fig:dataset_fig2}
\end{figure}

\textbf{Radiance and Density Grid Sampling from Instant-NGP NeRF:} NeRF-MAE reconstructs input masked NeRF volumes to pretrain standard 3D Swin Transformers. The first step in this process is to uniformly sample a radiance and density feature volume from a trained NeRF model~(Figure ~\ref{fig:dataset_fig2}). Querying a pretrained NeRF for radiance and density information on a regular grid allows~\textbf{1.} extracting an explicit representation that compactly captures the original 3D scene,~\textbf{2.} is invariant to the NeRF formulation used, and~\textbf{3.} opens up the possibility of utilizing existing deep-learning architectures developed for 3D tasks such as 3D RPN~(Section~\ref{subsec:downstream_tasks}) for detection and segmentation~\cite{hu2023nerf, liu2023instance}. To sample radiance and density information, we query each trained NeRF model from all the training cameras in the traceable scene volume and average the resulting output. 

Formally, Let $\mathcal{G} \in \mathbb{R}^{H \times W \times D \times 4}$ be a 4D grid representing the scene volume~(Figure~\ref{fig:dataset_fig2}~c), where, $H,W,D$ denote the height, width, and depth dimension. For a grid point $(i, j, k)$ on a spatial 3D grid, the 4 channel values for each grid point $(i, j, k)$ are the mean of the values obtained by this function $f(x, \theta)$ for all viewing directions. Mathematically, we can write this as:

\begingroup
\abovedisplayshortskip=-10pt
\begin{equation}
(r_{i, j, k}, g_{i, j, k}, b_{i, j, k}, \alpha_{i, j, k}) = \frac{1}{N} \sum_{\theta=1}^{N} f(x_{i, j, k}, \theta)
\end{equation}
\belowdisplayshortskip=-5pt
\endgroup
\looseness=-1

\noindent where $N$ is the number of training images, $\alpha$ can be obtained from volume density~($\sigma_{x}$) as $\alpha=1-\exp (-\sigma_i\|\boldsymbol{x}_i-\boldsymbol{x}_{i+1}\|)$ and ~$\boldsymbol{x}_i-\boldsymbol{x}_{i+1}\ =0.01$, is a small preset distance~\cite{hu2023nerf}. Similar to~\cite{hu2023nerf, liu2023instance}, we determine the traceable volume of the scene by enlarging the axis-aligned bounding box encapsulating all cameras and objects~(where available) in the scene. Compared to point-based pretraining approaches,~\cite{pang2022masked, zhang2022point} which carry information only at a surface-level and the distribution of information is relatively uneven~(see Fig.~\ref{fig:comparison_mae}), our approach makes use of dense information along all camera rays indicating that NeRF grids are similar to images in terms of information density; hence we show empirically that we are able to learn significantly better representations compared to other self-supervised pretraining baselines~(see. Section~\ref{sec:pretraining_3D_baelines}).

\subsection{NeRF Pretraining Datasets}
\label{subsec:nerf_pt_dataset}
A key component of our approach is employing a single model to pretrain strong 3D representations in a self-supervised manner from NeRFs on a large number of scenes with varying complexity and realism. As shown in Figure~\ref{fig:dataset_fig1}, we train and evaluate our representation in a fully self-supervised manner on the following dataset mix:

\noindent \textbf{Front3D}~\cite{fu20213d}: A large-scale dataset containing 18k rooms with 3D objects. We extend~\cite{hu2023nerf} to render 5x as many scenes, totaling 1.8k scenes comprising 425k images for our pretraining. The rendering time of this dataset is 6 days on 8 A100 GPUs.

\noindent \textbf{HM3D}~\cite{ramakrishnan2021hm3d}: A large-scale realistic dataset of 1000 scans. We render approximately 1.1k scenes comprising 1.2M images using Habitat~\cite{habitat19iccv}. Following~\cite{hong20233d}, we collect the data by randomly selecting 80 navigable points within each room's bounding box. At each navigable point, we rotate the agent 360$^{\circ}$ to render 12 frames with a $[10^{\circ},-10^{\circ}]$ elevation change. The rendering time of this dataset is around 2 days on 8 A100 GPUs.

\noindent \textbf{Hypersim}~\cite{roberts2021hypersim}: A synthetic dataset with real-world realism and 3D annotations. We use the rendered images~(250 scenes, 25k images) from~\cite{roberts2021hypersim} for our NeRF pretraining as they provide high-quality poses for reconstruction.

\noindent \textbf{ScanNet}~\cite{dai2017scannet}: A real-world dataset for indoor scenes. We keep this dataset for our cross-dataset transfer experiments and do not use this for pretraining. 90 out of the 1500 scenes are chosen, similar to~\cite{hu2023nerf} and we use the feature grid extracted from dense-depth-prior NeRF~\cite{roessle2022dense} for our downstream task experiments which further confirms the generalizability of our approach to various NeRF variants.

\looseness=-1
For all datasets, extensive cleaning of the bounding box as well as semantic annotations was performed based on manual filtering, class filtering, and size filtering (for further details, please see supplementary). 

\begin{figure*}[t!]
\centering
\includegraphics[width=1.0\textwidth]{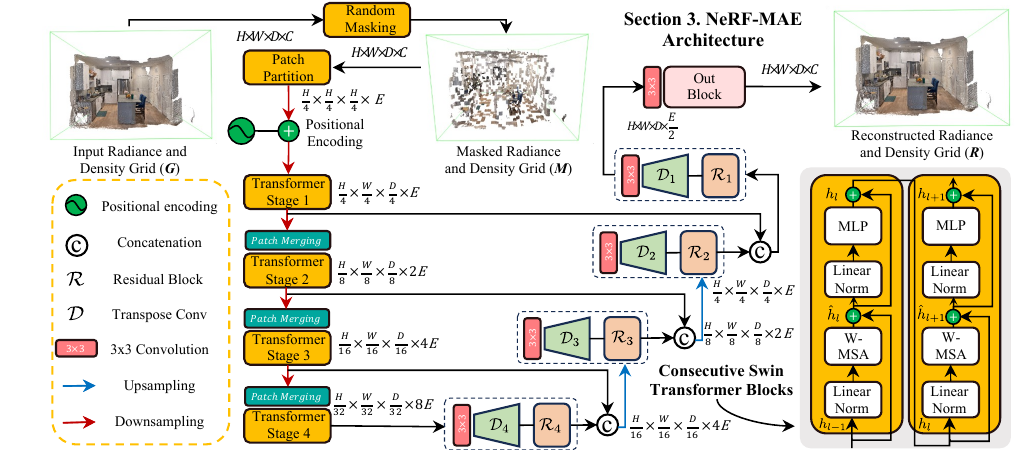}
\captionof{figure}{
\textbf{NeRF-MAE Architecture:} {Our method utilizes a U-Net~\cite{ronneberger2015u} style architecture employing Swin Transformers~\cite{liu2021swin} as the encoder to encode the radiance and density grid into meaningful multi-resolution low-level features, and transposed convolution layers at each stage with skip connections using residual blocks from the features of the encoder.
}}
\label{fig:architecture}
\end{figure*}

\subsection{Masked Pretraining NeRFs}
\label{subsec:masked_pretraining_nerf}
NeRF-MAE comprises a standard Swin Transformer~\cite{liu2021swin} encoder~($\mathbb{E}$) and transposed convolution decoders connected with skip connections from the encoded output at multiple feature resolutions of the transformer. Our architecture is summarized in Figure~\ref{fig:architecture}. The U-Net style architecture aims to reconstruct masked 3D radiance and opacity patches. In doing so, the network learns to understand the semantic and spatial structure of 3D scenes from dense 3D radiance and density grid~(obtained in Section~\ref{subsec:nerf_4d_grid}). Our U-Net~\cite{hatamizadeh2022unetr} style architecture for NeRF-pretraining is described below. 

\textbf{3D Transformer Encoder:}
The goal of the encoder is to encode the input 3D radiance and density grid~($\mathcal{G}$) into meaningful features at multiple resolutions which can be decoded and used for downstream tasks~(Section~\ref{subsec:downstream_tasks}). We employ the 3D version~\cite{yang2023swin3d, hu2023nerf} of the standard Swin Transformer~\cite{liu2021swin} which is obtained by replacing all the 2D operations of the SwinTransfomer with their 3D equivalent, as noted by~\cite{yang2023swin3d, hu2023nerf}; for example 3D convolutions and 3D patch merging. We build our 3D auto-encoder architecture based on standard Transformers since NeRF's radiance and density grid is similar to images in terms of information density and which can serve as a unified architecture for various 3D scene understanding tasks operating directly on NeRFs.

Considering the input to the standard 3D SwinTransformer~\cite{yang2023swin3d, hu2023nerf} to be the radiance and density grid~($\mathcal{G} \in \mathbb{R}^{H \times W \times D \times 4}$), we first mask out random patches of size~(P = $p\times p \times p$), where~$p=4$, with a masking ratio of~$m=0.75$ to get masked radiance and density grid~$\mathcal{M}$. Subsequently, we use~$P$ to divide the masked radiance and density grid into patches to create a series of 3D tokens of size~$\frac{H}{p}\times \frac{W}{p} \times \frac{D}{p}\times E$ using a 3D convolutional operation, where we merge patch partitioning and linear embedding of each patch into one single step. Next, we add 3D positional embedding to each patch, without a classification token, to assign each token to a unique representation. These tokens are further processed by four different stages of shifted window attention and patch merging, where the input patches are subdivided into non-overlapping windows of size~$W \times W \times W$, where~$W=4$, with local multi-head self-attention~(W-MSA-3D) and shifted multi-head self-attention~(SW-MSA-3D) performed for initial and subsequent layers in each block respectively~(see Fig.~\ref{fig:architecture}). Each Swin Transformer 3D block also contains a 2-layer~$MLP$ with GeLU activation as non-linearity in between following shifted window-based self-attention modules. A linear norm~($LN$) layer follows every~$\text{W-MSA-3D}$ and~$MLP$ with residual connection applied after each module. 

Due to the quadratic complexity in computing global self-attention for standard vision transformers, we employ shifted window attention~\cite{liu2021swin} by displacing the windows by~$[ \frac{W}{2}, \frac{W}{2}, \frac{W}{2}]$ volume every layer. A patch merging layer follows every transform block, except for the last one, which merges four volumetric patches, so a reduction by a factor of 2 in height, width, and depth of the input volume occurs. A Linear Norm~($LN$) and an MLP layer follow patch merging to increase the feature dimension by a factor of 2. We use an embedding dimension~$E=96$, Swin block depths of~$d=[2,2,18,2]$ and number of heads $[3,6,12,24]$ for each~$\text{W-MSA-3D}$ and~$\text{SW-MSA-3D}$ blocks. 

\begin{figure}[t]
    \centering
    \resizebox{1.0\linewidth}{!}{%
\includegraphics{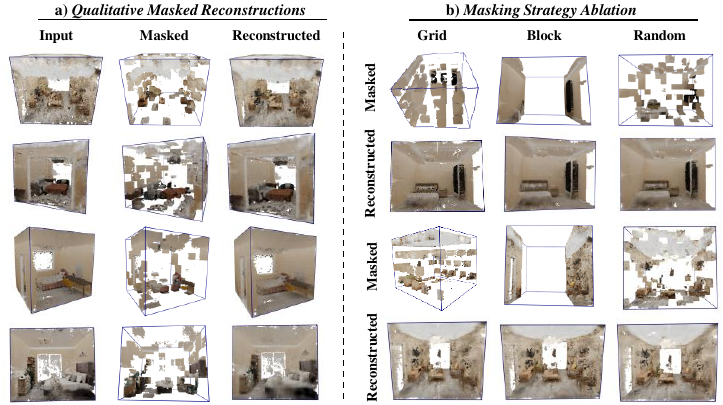}
    }
    \caption{~\textbf{Qualitative NeRF-MAE Reconstructions:}(\S~\ref{sec:qualitative_results})~\textbf{Left:} For each triplet, we show ground truth (left), masked radiance and density grid (middle) and our NeRF-MAE reconstruction overlayed with unmasked GT grid~(right). The masking ratio is 75\%, leaving only 250 patches out of 1000 patches.~\textbf{Right:} shows different masking strategies along with the reconstructed output.
    }
\label{fig:qualitative_recon}
\end{figure}%

\textbf{Reconstruction with Lightweight Decoders:} We utilize the feature pyramid~$\mathcal{F} = [f_{i} ... f_{n}]$ obtained after concatenating the resulting output of each SwinTransformer 3D stage to obtain the reconstructed radiance and density grid~($\mathcal{R}$), where~$i=0$ and $n=3$. The reconstructed grid is obtained by attaching lightweight transposed 3D convolution decoders~($\mathcal{D}_{i}$) to the feature outputs with a kernel size of 3 and adding a residual connection from the previous decoder, where~$d_{i} = Conv(d_{i-1} + \mathcal{D}(f_{i}))$ and~$\mathcal{D}_{0} = f_{3}$. The output of the final decoder block is fed into the residual block with~$3 \times 3 \times 3$ convolutional and a sigmoid activation function to predict a 4-channel reconstructed volume~$\mathcal{R} \in \mathbb{R}^{H \times W \times D \times 4}$. We use the same feature pyramid~$\mathcal{F}$ to output 3D object bounding boxes, voxel semantic labels, and super-resolution voxels with task-specific decoders, as described in Section~\ref{subsec:downstream_tasks}.

\textbf{Masked Radiance and Density Grid Pretraining Objective:}
The goal of masked voxel-grid pretraining as applied to Neural Radiance Field~(NeRF) is to encode semantic and spatial regions of interest into the network. We use mask volume reconstruction, similar to~\cite{he2021masked} for images. We enforce a faithful and accurate reconstruction of masked patches with a custom loss function suited to NeRF's unique formulation. Specifically, we employ a combination of opacity and photometric radiance reconstruction loss, both enforced at the volumetric level, where loss~$L_{recon} = \mathcal{L}_{\text{rad}} + \mathcal{L}_{\alpha}$ is defined as:

\begingroup
\abovedisplayshortskip=-10pt

\begin{equation*} \mathcal{L}_{\boldsymbol{rad}} = \frac{1}{K} \sum_{i=1}^{N} (\hat{\boldsymbol{y}}_{\text{rad}_i} - \boldsymbol{y}_{\text{rad}_i})^2 \quad \mathcal{L}_{\boldsymbol{\alpha}} = \frac{1}{M} \sum_{j=1}^{M} (\hat{\boldsymbol{y}}_{\alpha_j} - \boldsymbol{y}_{\alpha_j})^2 \end{equation*}
\belowdisplayshortskip=-12pt

\endgroup

\noindent where~$K$ is the number of voxels in the mask corresponding to patches where~$\alpha > \boldsymbol{x}_i-\boldsymbol{x}_{i+1}$. We use a small preset distance, ~$\boldsymbol{x}_i-\boldsymbol{x}_{i+1}\ = 0.01$.~$\hat{y}_{\text{rad}_i}$ is the predicted radiance at voxel~$i$ and~$y_{\text{rad}_i}$ is the target radiance at voxel~$i$.~$M$ denotes the number of voxels in the mask corresponding to removed 3D patches, ~$\hat{y}_{\alpha_j}$ is the predicted opacity at voxel,~$j$ and~$y_{\alpha_j}$ is the target opacity at voxel~$j$.
\section{Experiments}
\label{sec:experiments}

In this section, we aim to answer the following questions. 1) How well does our method \textit{compare with 3D pretraining baselines}? 2) Does NeRF-MAE~\textit{improve downstream 3D scene understanding tasks}? and 3) How well does~\textit{NeRF-MAE scale}? We also provide an \textit{ablation analysis} of which components most impact the performance.

\textbf{Baselines:} We compare 7 model variants across 3 different tasks to show the effectiveness of our method. 1)~\textbf{NeRF-RPN}~\cite{hu2023nerf}: State-of-the-art method that predicts 3D boxes directly in NeRFs. 2)~\textbf{Point-MAE}~\cite{pang2022masked}: Extends 2D MAE~\cite{he2021masked} to 3D pointclouds and improves classification and segmentation. 3)~\textbf{Point-M2AE}~\cite{zhang2022point}: Extends Point-MAE to add multiscale features for point cloud representation learning. 4)~\textbf{DepthContrast}~\cite{zhang_depth_contrast}: Uses depth to learn 3D representations in a contrastive manner by comparing transformations of a 3D point cloud. 5)~\textbf{Contrastive Scene Context}~\cite{hou2021exploring}: Leverages point correspondences along with spatial contexts to learn 3D representations. 6)~\textbf{ImVoxelNet}~\cite{rukhovich2022imvoxelnet}: A strong multi-view 3D detection baseline. 7)~\textbf{NeRF-MAE}: Our 3D representation learning approach utilizing posed 2D images as input and operating directly on scene's volumetric radiance and density grids obtained through NeRFs. 

\textbf{Datasets:}
We pretrain NeRF-MAE in a self-supervised manner on Front3D \cite{fu20213d}, HM3D~\cite{ramakrishnan2021hm3d} and Hypersim~\cite{roberts2021hypersim}. For cross-dataset transfer, we test on hold-out Scannet~\cite{dai2017scannet}, on splits provided by~\cite{hu2023nerf}. For further details, see Sec.~\ref{subsec:nerf_pt_dataset} and supplementary.

\textbf{Metrics:} We report~\textbf{3D PSNR} and~\textbf{3D MSE} to evaluate radiance grid reconstruction quality. We report Average Precision~\textbf{(AP)} along with~\textbf{Recall} at thresholds 25 and 50 to measure 3D OBB prediction success. We report mean intersection-over-union~\textbf{(mIOU)}, mean accuracy~\textbf{(mAcc)}, and total accuracy~\textbf{(Acc)} for the semantic voxel labeling task.

\captionsetup[table]{skip=6pt}
\begin{table}[t!]
\centering
\normalsize
\resizebox{\textwidth}{!}{
% \begin{tabular}{@{}lccccccccccc@{}}
\begin{tabular}{lccccccccccc}
\toprule
% \shline
& & \multicolumn{4}{c}{~\textit{Front3D~\cite{fu20213d} 3D OBB}} & \multicolumn{4}{c}{~\textit{ScanNet~\cite{dai2017scannet} 3D OBB}} \\
\cmidrule(lr){3-6} \cmidrule(lr){7-10}
 Approach & Aug.& \textbf{Recall@25}$\uparrow$ & \textbf{Recall@50}$\uparrow$ & \textbf{AP@25}$\uparrow$ & \textbf{AP@50}$\uparrow$ & \textbf{Recall@25}$\uparrow$ & \textbf{Recall@50}$\uparrow$ & \textbf{AP@25}$\uparrow$ & \textbf{AP@50}$\uparrow$ \\
% \hline
\midrule

%without FCAF-3D
NeRF-RPN~\cite{hu2023nerf}              &  \multicolumn{1}{c|}{-} &               0.961 &                      0.622 &                      0.780 &                     \multicolumn{1}{c|}{0.415} &                      0.891 &                      0.323 &                      0.491 &                      0.140 \\
ImVoxelNet~\cite{rukhovich2022imvoxelnet}            &  \multicolumn{1}{c|}{-}  &                 0.883 & \cellcolor{tabsecond}0.715 &  \cellcolor{tabfirst}\textbf{0.861} &  \multicolumn{1}{c|}{\cellcolor{tabfirst}\textbf{0.664}}  &                      0.517 &                      0.202 &                      0.373 &                      0.098 \\
% \hline
\midrule
NeRF-MAE (\textcolor{darkgreen}{F3D})  &  \multicolumn{1}{c|}{\textcolor{falured}{\xmark}}  &  \cellcolor{tabthird}0.962 &  \cellcolor{tabthird}0.675 &                      0.780 &                      \multicolumn{1}{c|}{0.543} &  \cellcolor{tabthird}0.897 &  \cellcolor{tabthird}0.361 &  \cellcolor{tabthird}0.510 &  \cellcolor{tabthird}0.145 \\
NeRF-MAE (\textcolor{darkgreen}{F3D})         &   \multicolumn{1}{c|}{\textcolor{teal}{\checkmark}} &  \cellcolor{tabsecond}\underline{0.963} &  \cellcolor{tabfirst}\underline{0.743} &  \cellcolor{tabthird}0.830 &  \multicolumn{1}{c|}{\cellcolor{tabthird}0.591}  & \cellcolor{tabsecond}\underline{0.905} & \cellcolor{tabsecond}\underline{0.391} & \cellcolor{tabsecond}\underline{0.543} & \cellcolor{tabsecond}\underline{0.155} \\
% \hline
\midrule
\textbf{NeRF-MAE (Ours)} & \multicolumn{1}{c|}{\textcolor{teal}{\checkmark}}& \cellcolor{tabfirst}\textbf{0.972} &  \cellcolor{tabfirst}\textbf{0.745} & \cellcolor{tabsecond}\underline{0.853} & \multicolumn{1}{c|}{\cellcolor{tabsecond}\underline{0.630}}  &  \cellcolor{tabfirst}\textbf{0.920} &  \cellcolor{tabfirst}\textbf{0.395} &  \cellcolor{tabfirst}\textbf{0.571} &  \cellcolor{tabfirst}\textbf{0.170} \\

\bottomrule
% \shline
\end{tabular}
}

\caption{\textbf{Quantitative comparisons with strong baselines on downstream 3D OBB prediction:} (\S~\ref{subsec:downstream_tasks}) showing our pretraining improves downstream performance on never-seen during pretraining cross-dataset~(ScanNet) and in-domain hold-out data~(Front3D). (\textbf{Ours}) indicate we use~\textcolor{darkgreen}{F3D}~\cite{fu20213d},~\textcolor{darkblue}{HM3D}~\cite{ramakrishnan2021hm3d} and~\textcolor{darkred}{HS}~\cite{roberts2021hypersim} for self-supervised pretraining. Aug. indicates augmentations done~\textit{only} during pretraining.}

\label{tab:downstream_3d_obb}
\end{table}

% NeRF-RPN~\cite{hu2023nerf} & - & 0.961 & 0.622 & 0.780 & 0.415
% &  0.891 & 0.323 & 0.491 & 0.140 \\
% \midrule
% NeRF-MAE (\textcolor{darkgreen}{F3D}) & \textcolor{falured}{\xmark}&  0.962 & 0.675 & 0.780 & 0.543
% &  0.897 & 0.361 & 0.510 & 0.145 \\

% NeRF-MAE (\textcolor{darkgreen}{F3D}) & \textcolor{teal}{\checkmark}&  0.963 & 0.743 & 0.830 & 0.59s1 &  0.905 & 0.391 & 0.543 & 0.155 \\
% \midrule
% % \hline
% NeRF-MAE (\textcolor{darkgreen}{F3D}+\textcolor{darkblue}{HM3D}+\textcolor{darkred}{HS}) & \textcolor{teal}{\checkmark}& \cellcolor{tabfirst}0.972 & \cellcolor{tabfirst}0.743 & \cellcolor{tabfirst}0.853 & \cellcolor{tabfirst}0.630 & \cellcolor{tabfirst}0.920 & \cellcolor{tabfirst}0.395 & \cellcolor{tabfirst} 0.571& \cellcolor{tabfirst}0.170 \\

% Note that we use hold-out scenes for transfer learning experiments.

% \zk{What does this last point mean? Hold out scenes from what?}\zk{Aren't there tons of works, both supervised and self/semi-supervised, that work on ScanNet/other datasets? How do we compare to them? Should we not put them in the table? It's hard to tell how our method compares to those works from this table. } \sz{Recall25-Front3D column is not highlighted correctly. Recall50 also has 2 baselines with equal results, but only one highlighted.}
% \vspace{-1cm}
\captionsetup[table]{skip=6pt}
\begin{table}[t!]
% \vspace{-2mm}
\begin{minipage}{.48\linewidth}
\scriptsize
\setlength{\tabcolsep}{0.4em}
\begin{tabular}{ccccccc}
\toprule
\textbf{Method} & \multicolumn{6}{c}{~\textit{3D OBB on ScanNet}} \\
\cmidrule(lr){1-1}\cmidrule(lr){2-7}
& \multicolumn{3}{c}{\textbf{Recall@50}$\uparrow$} & \multicolumn{3}{c}{\textbf{AP@50}$\uparrow$} \\
\midrule

& \textbf{S} & \textbf{PT} & \textbf{Diff.} & \textbf{S} & \textbf{PT} & \textbf{Diff.} \\
\midrule
\multicolumn{1}{c|}{DepthContrast~\cite{zhang_depth_contrast}}   & 7.5 & 11.7 & \multicolumn{1}{c|}{+4.2} & 2.4 & 4.1 & +1.7 \\

\multicolumn{1}{c|}{Scene Context~\cite{hou2021exploring}}  & 7.5 & 12.2 & \multicolumn{1}{c|}{+4.7} & 2.4 & 4.9 & +2.5 \\
\midrule
\multicolumn{1}{c|}{\textbf{NeRF-MAE (Ours)}}  & 32.3  & 39.5 & \multicolumn{1}{c|}{\textbf{+7.2}} & 14.0 & 17.0 & \textbf{+3.0}\\
\bottomrule

\end{tabular}
\end{minipage}

\hspace{1mm}
\begin{minipage}{.48\linewidth}
\scriptsize
\setlength{\tabcolsep}{0.4em}
\begin{tabular}{ccccccc}
\toprule
~\textbf{Method} & \multicolumn{6}{c}{~\textit{3D Segmentation on Front3D}} \\
\cmidrule(lr){1-1}\cmidrule(lr){2-7}
& \multicolumn{3}{c}{\textbf{mIOU}$\uparrow$} & \multicolumn{3}{c}{\textbf{mAcc}$\uparrow$} \\
\midrule

& \textbf{S} & \textbf{PT} & \textbf{Diff.} & \textbf{S} & \textbf{PT} & \textbf{Diff.} \\
\midrule
\multicolumn{1}{c|}{PointMAE~\cite{pang2022masked}}   & 15.1 & 17.3 & \multicolumn{1}{c|}{+2.2} & 20.9 & 22.8 & +1.9 \\

\multicolumn{1}{c|}{PointM2AE~\cite{zhang2022point}}  & 17.5 &  21.1& \multicolumn{1}{c|}{+3.6} & 23.4 & 27.6 & +4.2 \\
\midrule
\multicolumn{1}{c|}{\textbf{NeRF-MAE (Ours)}}  & 24.9  & 34.5 & \multicolumn{1}{c|}{\textbf{+9.6}} & 33.8 & 45.0 & \textbf{+11.2}\\
\bottomrule

\end{tabular}
\end{minipage}

% \centering
% \hfill\begin{minipage}{.45\linewidth}
% \scriptsize
% % \resizebox{0.45\textwidth}{!}{%
% \begin{tabular}{cccccccc}
% \toprule
% % \multicolumn{2}{c}{MAE Baselines} & \multicolumn{6}{c}{Metrics}\\
% Approach & Full Ann. & \multicolumn{6}{c}{Metrics on ScanNet test-set} \\
% \cmidrule(lr){1-2}\cmidrule(lr){3-8}
% % \midrule
% & & \multicolumn{3}{c}{\textbf{Recall50}} & \multicolumn{3}{c}{\textbf{AP50}} \\
% \midrule

% & & \textbf{S} & \textbf{PT} & \textbf{Diff.} & \textbf{S} & \textbf{PT} & \textbf{Diff.} \\

% \midrule
% DepthContrast~\cite{zhang_depth_contrast} &  \textcolor{teal}{\checkmark} & 7.5 & 11.7 & +4.2 & 2.4 & 4.1 & +1.7 \\

% Scene Context~\cite{hou2021exploring} &  \textcolor{teal}{\checkmark} & 7.5 & 12.2 & +4.7 & 2.4 & 4.9 & +2.5 \\
% \midrule
% NeRF-MAE (Ours) & \textcolor{falured}{\xmark} & 32.3  & 39.5 & \textbf{+7.2}& 14.0 & 17.0 & \textbf{+3.0}\\
% \bottomrule
% \end{tabular}
% \end{minipage}
% \vspace{1mm}
\caption{
\textbf{Quantitative comparison:}(\S~\ref{sec:pretraining_3D_baelines}) showing our approach's superior result compared to strong 3D representation learning and 3D MAE baselines using NeRF-rendered depth maps~(Left), and our NeRF-grid as input~(Right)~\cite{zhang_depth_contrast} and~\cite{hou2021exploring} uses VoteNet, which requires full annotations such as instance masks and mean class size.~\textbf{S} denotes starting from scratch and~\textbf{PT} denotes pretrained. Please refer to the supplementary for more details.
}
\label{tab:comparison_pretraining_baselines}
% \vspace{-8mm}
\end{table}

\subsection{Downstream tasks}
\label{subsec:downstream_tasks}

\textbf{3D Object Detection in NeRF}~\cite{hu2023nerf}: The task is to predict 3D OBBs. It entails regressing bounding box-offsets~$\boldsymbol{t}=\left(x_0, y_0, z_0, x_1, y_1, z_1, \Delta \alpha, \Delta \beta\right)$, objectness score~$c$ and a single objectness~$p$ for each voxel. The task is similar to anchor-free RPN from~\cite{hu2023nerf}.

\noindent \textbf{Voxel-grid Super-Resolution}: Given an input voxel grid~($V$) of resolution $160^3$, we predict an upsampled voxel-grid of higher resolutions, such as $384^3$. This can be denoted as~$V_{\text{upsample}}=g(V_{\text{input}}, \text{resolution})$, where~$g$ is a U-Net style encoder-decoder network. It is an important task for fast grid upsampling, since querying an implicit MLP for higher resolution is slow. 

\noindent \textbf{Semantic Voxel Labelling}: Given an input grid $G$ with dimensions $W \times  H\times D \times 4$, the task is to predict the class labels for each voxel. Every 3D voxel is assigned a class label, $S_{i,j,k}$ which are integers ranging from 1 to $n$, indicating the semantic category to which each voxel belongs. Mathematically, this can be represented as: $S_{i,j,k} \in \{1, 2, \ldots, n\}, \quad \text{for} \quad 1 \leq i \leq W, 1 \leq j \leq H, 1 \leq k \leq D$. Due to the diversity of objects in a scene, this task requires strong priors for generalization.

\subsection{Comparisons with strong baselines for improved downstream performance }
\label{sec:comparsons_downstream_tasks}

\textbf{3D Object Detection Downstream Task:} The results of our proposed method are summarized in Tables~\ref{tab:downstream_3d_obb} and~\ref{tab:amount_of_labelled_scenes}. We consistently outperform the state-of-the-art baseline methods on 3D object detection in NeRFs on both Front3D and ScanNet scenes. Among our variants, our method pretrained on multiple data sources~(Table~\ref{tab:downstream_3d_obb},~row 5) and with augmentations enabled during pretraining performs the best. Specifically, NeRF-MAE shows superior performance on the unseen test-set by achieving an AP50 of 63\% and a Recall50 of 74.3\%, hence demonstrating an absolute improvement of 21.5\% in AP50 and 12.1\% in Recall50 on the Front-3D dataset. Our method also achieves superior performance on ScanNet which is never seen during pretraining. We achieved an AP50 of 17\% and a Recall50 of 39.5\%, hence demonstrating an absolute performance improvement of 3\% AP50 and 7.2\% on Recall50 over the best-performing baseline.

\begin{table*}[t!]
\centering
\scriptsize
\resizebox{\textwidth}{!}{
\begin{tabular}{@{}ccccccccccc@{}}
\toprule
% \shline
&  & \multicolumn{4}{c}{~\textit{MAE pre-trained encoder (\textbf{Ours})}} & \multicolumn{4}{c}{~\textit{NeRF-RPN~\cite{hu2023nerf} Start from scratch}} \\
\cmidrule(lr){3-6} \cmidrule(lr){7-10}
 \#Labelled & \#Scenes & \textbf{Recall@25}$\uparrow$ & \textbf{Recall@50}$\uparrow$ & \textbf{AP@25}$\uparrow$ & \textbf{AP@50}$\uparrow$ & \textbf{Recall@25}$\uparrow$ & \textbf{Recall@50}$\uparrow$ & \textbf{AP@25}$\uparrow$ & \textbf{AP@50}$\uparrow$ \\
\midrule
\multicolumn{1}{c|}{10\%} & \multicolumn{1}{c|}{12}& 0.93 & 0.41 & 0.52 & \multicolumn{1}{c|}{0.18} & 0.91 & 0.35 & 0.49 & 0.15 \\
\multicolumn{1}{c|}{25\%} & \multicolumn{1}{c|}{30} & 0.94 & 0.51 & 0.67 & \multicolumn{1}{c|}{0.36} & 0.94 & 0.49 & 0.67 & 0.29 \\
\multicolumn{1}{c|}{50\%} & \multicolumn{1}{c|}{61} & 0.95 & 0.60 & 0.74 & \multicolumn{1}{c|}{0.42} & 0.95 & 0.58 & 0.71 & 0.30\\
\multicolumn{1}{c|}{100\%} & \multicolumn{1}{c|}{122} & \cellcolor{tabfirst}\textbf{0.96} & \cellcolor{tabfirst}\textbf{0.67} & \cellcolor{tabfirst}\textbf{0.79} & \multicolumn{1}{c|}{\cellcolor{tabfirst}\textbf{0.54}}  & \cellcolor{tabfirst}\textbf{0.96} & \cellcolor{tabfirst}\textbf{0.62} & \cellcolor{tabfirst}\textbf{0.78} & \cellcolor{tabfirst}\textbf{0.41} \\
\bottomrule
% \shline
\end{tabular}
}
\caption{\textbf{Effect of amount of 3D labeled scenes on downstream 3D object detection performance:}(\textbf{\S}~\ref{subsec:scaling_laws_and_ablation}) showing our approach's superior results compared to strong baseline. The FPN network weights in both cases are initialized from scratch.}
% \vspace{-0.2cm}
\label{tab:amount_of_labelled_scenes}
\end{table*}
% \vspace{-1cm}
\captionsetup[table]{skip=2pt}
\begin{table}[t!]
% \vspace{-2mm}
\begin{minipage}{.46\linewidth}
\scriptsize
\setlength{\tabcolsep}{0.4em}
\centering

\begin{tabular}{ccccc}

\toprule
\multicolumn{5}{c}{~\textit{Super-Resolution Downstream Task}} \\
\midrule
& \multicolumn{2}{c}{$256^{3}$} & \multicolumn{2}{c}{$384^{3}$}\\
\cmidrule(lr){1-2}
\cmidrule(lr){2-3} \cmidrule(lr){4-5}
 Method & \textbf{PSNR}$\uparrow$& \textbf{MSE} $\downarrow$&  \textbf{PSNR}$\uparrow$& \textbf{MSE} $\downarrow$ \\ 
\midrule

 \multicolumn{1}{c|}{NeRF-RPN~\cite{hu2023nerf}}  & 16.25 &\multicolumn{1}{c|}{0.024} & 16.08 &0.025 \\
\multicolumn{1}{c|}{\textbf{Ours}}   &\cellcolor{tabfirst}\textbf{17.27} &\multicolumn{1}{c|}{\cellcolor{tabfirst}\textbf{0.019}} & \cellcolor{tabfirst}\textbf{17.34} &\cellcolor{tabfirst}\textbf{0.019} \\
\bottomrule
\end{tabular}

% \caption{\textbf{Voxel super-resolution quantitative results:} showing superior performance of our method vs NeRF-RPN~\cite{hu2023nerf}. * denotes we add a task-specific head to NeRF-RPN's Swin-S encoder.}

% \vspace{-12pt}
% \label{tab:voxel_super_resolution}
\end{minipage}
% \end{table}
\hspace{1mm}
\begin{minipage}{.54\linewidth}
\scriptsize
\setlength{\tabcolsep}{0.4em}
\centering

\begin{tabular}{ccccc}
\toprule

\multicolumn{5}{c}{~\textit{Semantic Voxel-Labelling Downstream Task}} \\
\midrule
& \multicolumn{2}{c}{Front3D~\cite{fu20213d}} & \multicolumn{2}{c}{HM3D~\cite{ramakrishnan2021hm3d}} \\
\cmidrule(lr){2-3} \cmidrule(lr){4-5}
Metrics & NeRF-RPN~\cite{hu2023nerf} & ~\textbf{Ours}& NeRF-RPN~\cite{hu2023nerf} & ~\textbf{Ours} \\
 % \midrule
\hline
\multicolumn{1}{c|}{\textbf{mIOU}$\uparrow$}  & 0.249 & \multicolumn{1}{c|}{\cellcolor{tabfirst}\textbf{0.345}}  & 0.109 & \cellcolor{tabfirst}\textbf{0.186} \\
\multicolumn{1}{c|}{\textbf{mAcc}$\uparrow$}  & 0.338 & \multicolumn{1}{c|}{\cellcolor{tabfirst}\textbf{0.450}}  & 0.160 & \cellcolor{tabfirst}\textbf{0.264} \\
\multicolumn{1}{c|}{\textbf{Acc}$\uparrow$}  & 0.730 & \multicolumn{1}{c|}{\cellcolor{tabfirst}\textbf{0.810}}  & 0.510 & \cellcolor{tabfirst}\textbf{0.581} \\
\bottomrule
% \end{tabular}
\end{tabular}
% \caption{\textbf{Semantic voxel-labelling results:} showing superior performance of our method vs NeRF-RPN~\cite{hu2023nerf}. * denotes we add a task-specific head to NeRF-RPN's Swin-S encoder.}
% \vspace{-14pt}
% \label{tab:voxel_labelling}
\end{minipage}
% \vspace{1mm}
\caption{
\textbf{Voxel semantic-labelling and super-resolution quantitative results:}(\textbf{\S}~\ref{sec:comparsons_downstream_tasks}) showing superior performance of our method vs NeRF-RPN~\cite{hu2023nerf}.
}
% \vspace{-8mm}
\label{tab:comparison_downsteam_voxel_superresolution}
\end{table}

\noindent \textbf{Semantic Voxel-labelling and Voxel-Super Resolution:} We further test our approach's ability on two challenging downstream tasks~i.e. semantic voxel-labelling and voxel super-resolution. The results are summarized in Tab.~\ref{tab:comparison_downsteam_voxel_superresolution}. The results clearly show our proposed approach consistently outperforms the strong baseline on all metrics which further indicates our networks' ability to learn good representation for the challenging downstream 3D task. Specifically, we achieve 81\% Acc, 45\% mAcc, and 34.5\% mIOU on voxel-labeling task which equals an absolute performance improvement of 6.8\% Acc, 12.9\% mAcc, and 9.8\% mIOU metric against NeRF-RPN~\cite{hu2023nerf} on Front3D. Our method also achieves better 3D PSNR and low MSE metrics~(absolute improvement of 1.02 for~$256^3$ and 1.259 for~$384^3$ grid show) on voxel-super resolution, showing our network learns strong representation for this dense task.

\subsection{Comparisons with strong 3D representation learning baselines}
\label{sec:pretraining_3D_baelines}

Table~\ref{tab:comparison_pretraining_baselines} presents the quantitative evaluations of our approach's performance in comparison with strong 3D representation learning baselines. NeRF-MAE consistently outperforms all competing methods with a clear margin. For a fair comparison, we provide NeRF-rendered depth maps to DepthContrast~\cite{zhang_depth_contrast} and SceneContext~\cite{hou2021exploring} while providing our NeRF-grid to Point-MAE~\cite{pang2022masked} and PointM2AE~\cite{zhang2022point}. Specifically, we achieve +7.2\% Recall50 improvement and +3\% AP50 improvement, hence demonstrating an absolute improvement number of +2.5\% Recall50 and +0.5\% AP50 on Scannet 3D OBB prediction task. We also achieve a +9.6\% mIOU improvement and +11.2\% mAcc improvement on Front3D voxel labelling, hence demonstrating an absolute improvement number of +6\% mIOU and +7\% mAcc over the best competing 3D pretraining baseline. This shows our method learns better representations by utilizing a dense volumetric NeRF-grid as well as an opacity-aware reconstruction objective.

\begin{figure}[t]
    \centering
    \resizebox{0.90\linewidth}{!}{%
\includegraphics{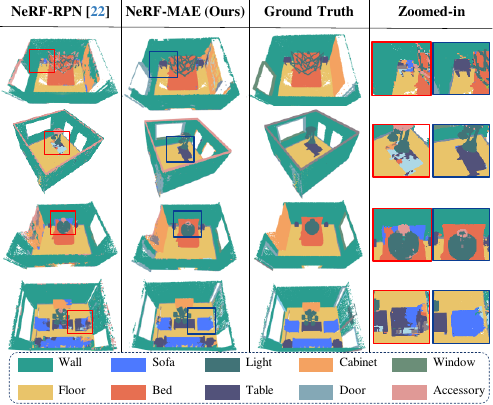}
    }
    \caption{\textbf{Qualitative voxel-labelling downstream task generalization comparison:}(\textbf{\S}~\ref{sec:qualitative_results}) showing our approach's superior results compared with NeRF-RPN~\cite{hu2023nerf}.
    }
    \label{fig:qualitative_recon_5}
\end{figure}%

\begin{figure}[t]
    \centering
    \resizebox{0.98\linewidth}{!}{%
\includegraphics{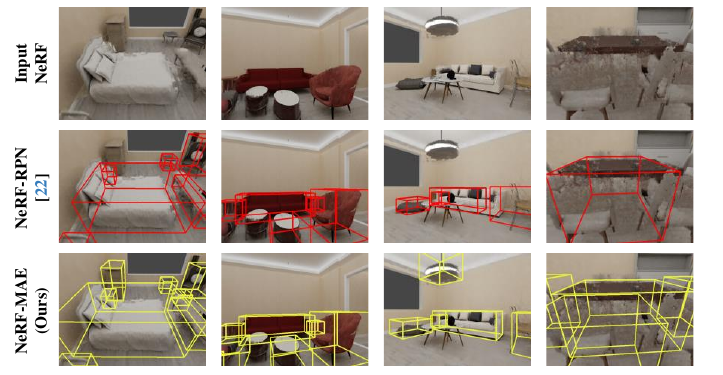}
    }
    \caption{\textbf{Qualitative 3D OBB prediction downstream task generalization comparison:}(\textbf{\S}~\ref{sec:qualitative_results}) showing our approach's superior results compared with NeRF-RPN~\cite{hu2023nerf}.}
    \label{fig:qualitative_recon_3}
\end{figure}%

\subsection{Scaling Performance and Ablation Analysis}
\label{subsec:scaling_laws_and_ablation}

\looseness=-1
We analyze our network's ability to learn useful 3D representations by studying how the amount and quality of pretraining data affects downstream performance. We summarize our results in Figure~\ref{fig:quantitative_results_scaling_laws}. To study scaling laws, we select Front3D dataset and use 10\%, 25\%, 50\%, and 100\% of the scenes for pretraining. Our test-set MSE and AP50 metrics clearly show our network's ability to improve with more unlabeled data on the reconstruction and transfer learning tasks. Specifically, we achieved an absolute AP50 improvement of 10\% when adding 10x the number of scenes for pretraining~(1515 vs 151 scenes), emphasizing our network is able to learn better on more unlabelled data. To understand how the quality of NeRFs during the pretraining stage affects downstream performance, we train the input NeRFs to varying levels of accuracy, providing all 100\% of the scenes at various PSNR values. Our results show that we can learn better representations, achieving downstream AP25 improvement of 36\% when adding higher quality NeRFs to our pretraining~(18 vs 28 2D PSNR).

\noindent \textbf{Amount of unlabeled scenes on downstream 3D OBB task:} We evaluate our method on the 3D OBB prediction task by limiting the available scenes for transfer learning.  As shown in Tab.~\ref{tab:amount_of_labelled_scenes}, our method surpasses the baseline across all scenarios, notably achieving a 42\% AP50 with less than half the data requirement of the strong baseline; further highlighting our approach's ability to enhance performance in data-limited settings.

\section{Qualitative Results}
\label{sec:qualitative_results}
\looseness=-1
We qualitatively analyze the performance of NeRF-MAE on reconstruction, 3D OBB prediction, and voxel-labeling tasks. As shown in Fig.~\ref{fig:qualitative_recon}, our approach outputs plausible reconstructions showing our network's ability to find spatial relationships in the input data, which is useful for learning good representations. We further show our networks' superior 3D OBB prediction performance in Figure~\ref{fig:qualitative_recon_3}. Our approach reconstructs accurate bounding boxes where the state-of-the-art baseline struggles with incomplete or erroneous detections. Finally, we show our networks' superior semantic segmentation performance in Figure~\ref{fig:qualitative_recon_5}. This result further emphasizes our network learns good representations for a challenging 3D downstream task.

\section{Conclusion}
\label{sec:conclusion}
Neural Fields have shown emergent properties going beyond reconstruction; hence making them useful for inferring semantics, geometry, and dynamics. In this work, we use masked auto-encoders to study scaling NeRFs for their self-supervised pretraining. In essence, we use a single standard Transformer model to learn useful 3D representations from NeRF which we show to be highly effective for 3D transfer learning. To pretrain and evaluate our representation, we curate a large-scale dataset totaling 1.8M+ images and 3,600+ scenes from 4 different sources. Our approach, NeRF-MAE significantly outperforms self-supervised 3D pretraining as well as NeRF scene understanding baselines on a variety of downstream tasks. As it requires readily available and cheaper to obtain posed-RGB data and with a strong empirical quantitative performance over baselines, we hope this avenue is a potential future direction of 3D pertaining. \\

\noindent \textbf{Acknowledgements}: We thank Greg Shakhnarovich and Katherine Liu for valuable feedback.

% \section*{Acknowledgements}
% We thank Greg Shakhnarovich and Katherine Liu for their valuable feedback.

\bibliographystyle{splncs04}
\bibliography{main}

\clearpage
\begin{center}
\vspace{0.2cm}
    {\Large \textbf{ NeRF-MAE: Masked AutoEncoders for Self-Supervised 3D Representation Learning for Neural Radiance Fields \\[0.5em] --- Supplementary Material ---}}
\end{center}

\setcounter{page}{1}
\setcounter{section}{0} % Reset section
\renewcommand{\thesection}{\Alph{section}} % Section counter format: A, B, C, ...
% \makeatletter

This supplementary material is divided into 5 sections. First, we provide details regarding network architecture and number of parameters in Section~\ref{sec:architecture_details}. Second, we provide an overview of implementation details in Section~\ref{sec:implementation_details}. This is followed by a discussion on additional experimental analysis in Section~\ref{sec:add_experiments}. We provide an overview of pretraining datasets in Section~\ref{sec:pretraining_datasets} and discuss compute resources required to pretrain our representation in Section~\ref{sec:compute_resources}. Lastly, we discuss limitations and future work in Section~\ref{sec:future_work}.

\section{Network Architecture Details}
\label{sec:architecture_details}
In this section, we provide more details about our architectural design. We use PyTorch~\cite{paszke2019pytorch} for implementation and training our approach. We described in detail the encoder-decoder U-Net style architecture for pretraining in Section~\ref{subsec:masked_pretraining_nerf}~(in the main paper). We use a constant W, D, H dimension of 160 for all of our datasets, a patch size $p=4$, and mask out randomly 75\% of the patches. We use 4 Swin Transformer blocks and the spatial feature size of the last block is $5\times5\times5$ with the channel dimension being 768. Consequently, we upsample features with skip connection decoders and use 4 decoders and a final convolution block to restore the original dimension i.e. $B\times4\times160\times160\times160$. Below we specify the specifications and architectures of our task-specific decoders. 

\subsection{Task-specific Downstream 3D Heads}

Our task-specific downstream heads are attached to the Swin-Transformer encoder. The goal of task-specific heads is to output dedicated meaningful outputs to regress the useful 3D information such as 3D OBBs and semantic labels for each voxel. \\

\noindent\textbf{3D OBB Prediction:} Our 3D OBB prediction head is similar to NeRF-RPN~\cite{hu2023nerf} except that for all of our transfer learning experiments and comparisons, we start with warm-started encoder weights via our large-scale self-supervised pretraining, whereas NeRF-RPN weights are started from scratch. Starting from scratch here denotes that the network weights were not pretrained on any other dataset but rather NeRF-RPN initializes the weights of linear layers with truncated normal distribution with a standard deviation of 0.02. We use the author's~\cite {hu2023nerf} official implementation for all the comparisons. Specifically, we utilize an anchor-free RPN head from~\cite{hu2023nerf} which is based on the FCOS detector lifted to the 3D domain via replacing 2D convolutions with their 3D counterparts. We employ a combination of focal loss, IOU loss, and binary cross-entropy loss for downstream transfer learning task similar to~\cite{hu2023nerf}. \\

\noindent\textbf{3D Semantic Voxel-Labelling:} The task of semantic voxel labeling is described in detail in Section~\textcolor{red}{~\ref{subsec:downstream_tasks}}~(in the main paper). Specifically, we use the same decoders with skip connections i.e.~$\mathcal{D}_{i}$, as described in Section~\ref{subsec:masked_pretraining_nerf}~(in the main paper) where~$i$ ranges from 2 to 4. We replace~$\mathcal{D}_{1}$ by incorporating an additional skip connection with the encoded output from the input grid~($\mathcal{G}$). This is followed by a final convolutional output block to output semantic volume~($\mathcal{G}_{S} \in \mathbb{R}^{H \times W \times D \times C}$) of spatial dimension $W,H,D$ where $W=H=D=160$ and $C=18$ for the Front3D dataset and 20 for the Habitat-Matterport3D dataset. We use a weighted masked cross-entropy loss, balanced by the inverse log propensity of each class, where propensity is defined to be the frequency of each class in the training dataset.  \\

\noindent\textbf{Voxel Super-Resolution:} The task of voxel super-resolution is described in detail in Section~\ref{subsec:downstream_tasks}~(in the main paper). Specifically, we replace convolution decoders with four 3D convolution layers with instance norm and ReLU activation followed by an upsampling by a factor of 2. This is followed by a final upsampling of 1.6 and 2.4 depending on the output resolution size of~$256^3$ and~$384^{3}$ and a final 3D convolution. Similar to the original reconstruction loss~(Section~\ref{subsec:masked_pretraining_nerf}~(in the main paper)), we employ a masked color reconstruction loss enforced with the ground-truth upsampled grid. 

\begin{figure}[t]
\includegraphics[width=1.0\textwidth]{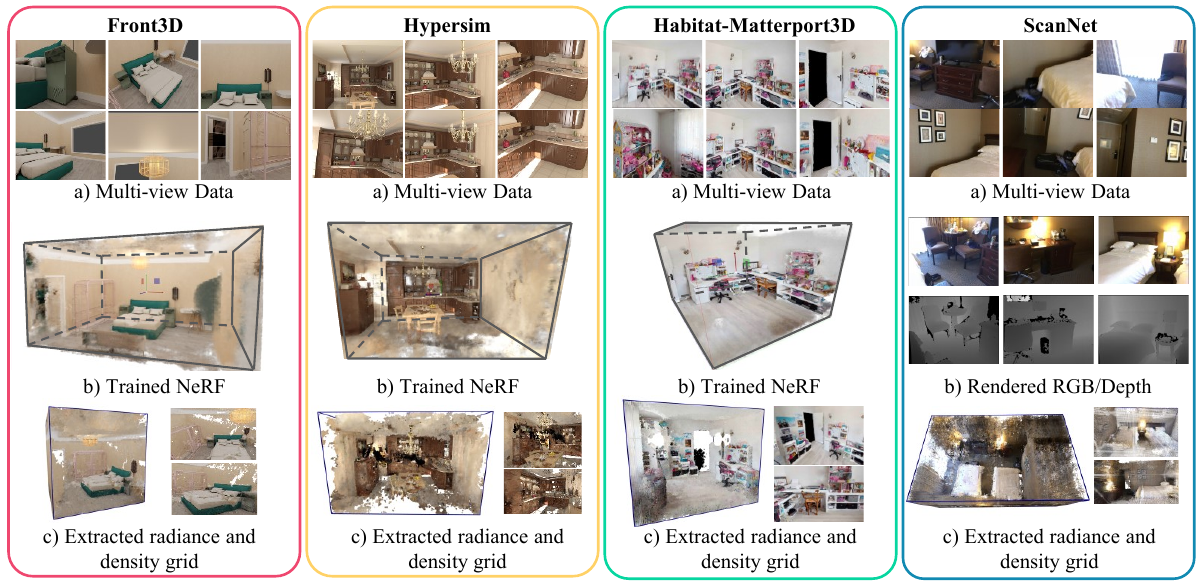}

\caption{\textbf{Data preparation for pretraining:} showing multi-view images, trained NeRF and extracted radiance and density grid for Instant-NGP~\cite{muller2022instant} trained NeRFs for Front3D~\cite{fu20213d}, HyperSim~\cite{roberts2021hypersim} and Habitat-Matterport3D~\cite{ramakrishnan2021hm3d}. The right column shows multi-view images, rendered RGB/depth images, and extracted radiance and density grid for ScanNet~\cite{dai2017scannet} dataset, which is trained using dense depth prior NeRF~\cite{roessle2022dense}. Note that we do not pre-train our representation on ScanNet, rather use this dataset for our cross-dataset transfer experiment for the downstream 3D OBB prediction task.}
\label{fig:pretraining_datasets}
\end{figure}

\begin{figure}[t!]
\centering
\includegraphics[width=1.0\textwidth]{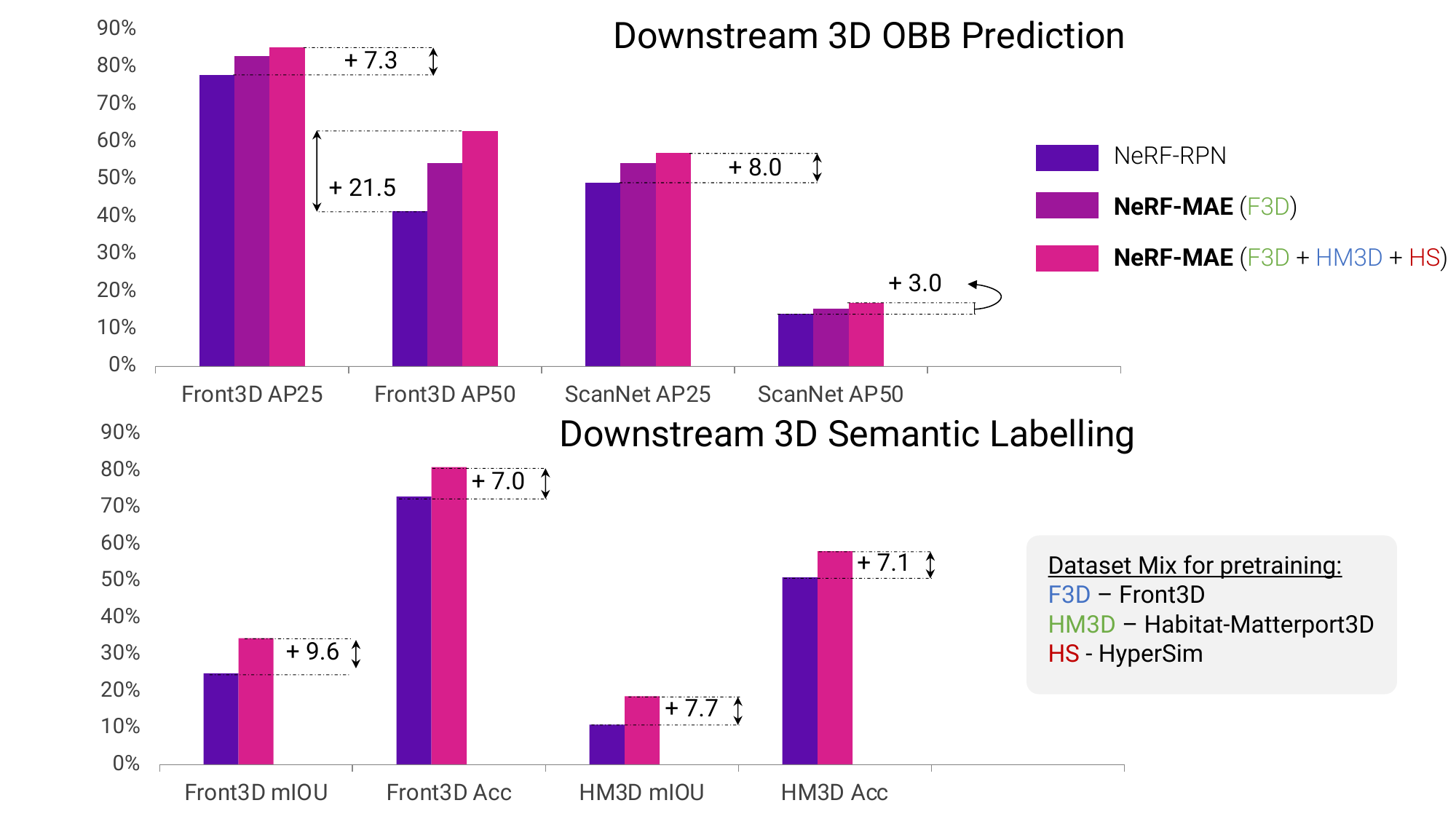}
\captionof{figure}{
\textbf{Quantitative Comparison:} We show further quantitative analysis of results reported in Table~\ref{tab:downstream_3d_obb} and Table~\ref{tab:comparison_downsteam_voxel_superresolution} in our main paper. Specifically, we highlight improvement numbers for AP25 and AP50 on ScanNet~\cite{dai2017scannet} and Front3D~\cite{fu20213d} and mIOU and Acc for HM3D~\cite{ramakrishnan2021hm3d} and Front3D~\cite{fu20213d} over the state-of-the-art baseline. Furthermore, we also show that increasing unlabelled posed 2D data from different sources improves performance on 3D OBB prediction downstream task. 
}
\label{fig:comparison_chart2}
\end{figure}

\subsection{Parameters}
NeRF-MAE efficiently trains on diverse scenes using a 70 million parameter network, highlighting its capability to handle complex and varied training data. Our Swin-S Transformer encoder architecture as well as the anchor-free RPN head remains the same as NeRF-RPN~\cite{hu2023nerf} for a fair comparison on the downstream 3D OBB prediction task. For reconstruction, we employ lightweight decoders~(Section~\ref{subsec:masked_pretraining_nerf} in the main paper) since for our transfer learning experiments, we eventually discard the decoders and utilize the feature pyramids for downstream 3D tasks as described above.

\section{Implementation Details}
\label{sec:implementation_details}

To pre-train NeRF-MAE, we utilize the Swin-S backbone of the SwinTransformer. As detailed in Section~\ref{subsec:masked_pretraining_nerf}~(in the main paper), we utilize four swin blocks of depth 2, 2, 18, and 2 with the number of heads in each block equalling 3, 6, 12, and 24. We pre-train both encoder and decoder networks for 1200 epochs with the Adam optimizer and a maximum learning rate of~$3e^{-4}$, a weight decay of~$1e^{-3}$ with a one-cycle learning rate scheduling. We employ a batch size of 32 for our pretraining. We additionally utilize online data augmentation during both pretraining and transfer learning stages. We employ random flip, rotation, and scaling augmentation with a probability of 50\%. Data augmentation is shown to improve our pretraining, as described in Section~\ref{sec:experiments}. For transfer learning, we train both our network and the baseline NeRF-RPN~\cite{hu2023nerf} for 1000 epochs for 3D OBB prediction and 500 epochs for voxel super-resolution with the same learning rate and strategy as our pretraining stage since we empirically determined that changing the learning rate and strategy results in decreased performance for both our method and the baseline. We additionally utilize a maximum normal gradient clip value of 0.1 and fine-tune the network until convergence. We determine transfer learning convergence on the hold-out validation set and employ early stopping based on AP50 metrics for 3D OBB prediction, accuracy metric for 3D semantic voxel labeling, and 3D PSNR metric for voxel super-resolution experiments. For transfer learning experiments, we use a batch size of 8.

\textbf{Pretraining and Fine-tuning Weights:} In the spirit of the original 2D MAE work~\cite{he2021masked}, the pretraining weights of the model are general-purpose, containing the scene understanding knowledge such as semantic and spatial structure. Finetuning allows the model weights to be task-specific and aligns the model with the task using lightweight decoders. We will further include a detailed discussion on this topic in our paper.

% \vspace{-0.1mm}
\captionsetup[table]{skip=2pt}
\begin{table}[t!]
% \vspace{-2mm}
\begin{minipage}{.56\linewidth}
\scriptsize
\setlength{\tabcolsep}{0.7em}
\centering

% \begin{table}[t!]
% \centering
% \setlength{\tabcolsep}{2.5pt}
% \scriptsize
\begin{tabular}{ccccc}

\toprule

& \multicolumn{2}{c}{$256^{3}$} & \multicolumn{2}{c}{$384^{3}$}\\
\cmidrule(lr){1-2}
\cmidrule(lr){2-3} \cmidrule(lr){4-5}
Method & \textbf{PSNR}$\uparrow$& \textbf{MSE} $\downarrow$&  \textbf{PSNR}$\uparrow$& \textbf{MSE} $\downarrow$ \\ 
\midrule
% \hline 

\multicolumn{1}{c|}{NeRF-RPN*~\cite{hu2023nerf}}  & 14.70 & \multicolumn{1}{c|}{0.034}  & 14.65 &0.035 \\
\multicolumn{1}{c|}{\textbf{NeRF-MAE(Ours)}}  &\cellcolor{tabfirst}\textbf{15.01} &\multicolumn{1}{c|}{\cellcolor{tabfirst}\textbf{0.032}} 
  &\cellcolor{tabfirst}\textbf{15.20} & \cellcolor{tabfirst}\cellcolor{tabfirst}\textbf{0.030}  \\
\bottomrule
\end{tabular}
% \caption{\textbf{Voxel super-resolution quantitative results:} for HM3D~\cite{ramakrishnan2021hm3d} test-set showing superior performance of our method vs NeRF-RPN~\cite{hu2023nerf}. * denotes that we add a task-specific head to NeRF-RPN's Swin-S encoder.}
% \label{tab:voxel_super_resolution_hm3d}
% \end{table}

\end{minipage}
\hspace{1mm}
\begin{minipage}{.42\linewidth}
\scriptsize
\setlength{\tabcolsep}{0.6em}
\centering

% \begin{table}[t!]
% \setlength{\tabcolsep}{2.5pt}
% \centering
% \footnotesize

% \resizebox{0.45\textwidth}{!}{%
\begin{tabular}{ccc}

\toprule
\multicolumn{3}{c}{\textit{3D OBB on HM3D~\cite{ramakrishnan2021hm3d}}} \\
\midrule
Method & \textbf{AP@25}$\uparrow$  & \textbf{Recall@50}$\uparrow$ \\
\midrule
\multicolumn{1}{c|}{NeRF-RPN~\cite{hu2023nerf}}  & 0.30  & 0.31\\
\multicolumn{1}{c|}{\textbf{NeRF-MAE (Ours)}}  & \cellcolor{tabfirst}\textbf{0.37} & \cellcolor{tabfirst}\textbf{0.40} \\
\bottomrule
\end{tabular}
% }

% \caption{\textbf{3D OBB prediction quantitative result} for HM3D~\cite{ramakrishnan2021hm3d} test-set showing our approach's superior performance compared with NeRF-RPN~\cite{hu2023nerf}}.
% \vspace{-10pt}
% \label{tab:hm3d_obb}
% \end{table}

\end{minipage}
% \vspace{1mm}
\caption{\textbf{Voxel super-resolution and 3D OBB prediction quantitative results} for HM3D~\cite{ramakrishnan2021hm3d} test-set showing superior performance of our method vs NeRF-RPN ~\cite{hu2023nerf}. These results further confirm the findings of Tables~\ref{tab:downstream_3d_obb} and~\ref{tab:comparison_downsteam_voxel_superresolution} in our main paper that our pretraining helps improve downstream tasks on various datasets.
}
% \vspace{-2mm}
\label{tab:comparison_downsteam_hm3d}
\end{table}
% \vspace{-0.1mm}
\captionsetup[table]{skip=2pt}
\begin{table}[t!]
% \vspace{-2mm}
\begin{minipage}{.54\linewidth}
\scriptsize
\setlength{\tabcolsep}{0.25em}
\centering

% \begin{table}[t!]
% \centering
% \scriptsize
% \resizebox{0.45\textwidth}{!}{%
\begin{tabular}{cccccc}
\toprule
\multicolumn{6}{c}{\textit{Pretraining scenes on downstream performance}} \\
\midrule
\multicolumn{2}{c}{Pretraining Scenes} & \multicolumn{4}{c}{Metrics}\\
\cmidrule(lr){1-2}\cmidrule(lr){3-6}
\% Scenes& \multicolumn{1}{c|}{\# Scenes} & \textbf{Recall@25}$\uparrow$ & \textbf{Recall@50}$\uparrow$ & \textbf{AP@25}$\uparrow$ & \textbf{AP@50}$\uparrow$ \\
\midrule
10\% & \multicolumn{1}{c|}{151}  & 0.94 & 0.61 & 0.74 & 0.45 \\
25\% & \multicolumn{1}{c|}{378}  & 0.95 & 0.62 & 0.76 & 0.46 \\
50\% & \multicolumn{1}{c|}{757}  & 0.95 & 0.66 & 0.77 & 0.51 \\
100\% & \multicolumn{1}{c|}{1515} & \cellcolor{tabfirst}\textbf{0.96} & \cellcolor{tabfirst}\textbf{0.67} & \cellcolor{tabfirst}\textbf{0.79} & \cellcolor{tabfirst}\textbf{0.54} \\
\bottomrule
\end{tabular}
% }

% \caption{\textbf{Effect of percent of pretraining scenes on downstream 3D object detection performance} showing more unlabelled data helps learn better representation for downstream 3D tasks.}
% \vspace{-5pt}
% \label{tab:effect_pretraining}
% \end{table}

\end{minipage}
\hspace{1mm}
\begin{minipage}{.46\linewidth}
\scriptsize
\setlength{\tabcolsep}{0.25em}
\centering

% \begin{table}[t!]
% \centering
% \scriptsize
% \resizebox{0.45\textwidth}{!}{%
\begin{tabular}{ccccc}
\toprule
\multicolumn{5}{c}{\textit{Pretraining scenes on reconstruction quality}} \\
\midrule \\
[-1.5em]
\multicolumn{3}{c}{Pretraining Scenes} & \multicolumn{2}{c}{Metrics}\\
\cmidrule(lr){1-3}\cmidrule(lr){4-5}
\textbf{\% Scenes} & \textbf{\# Scenes} & ~\textbf{Aug.} &\textbf{3D PSNR} $\uparrow$ & \textbf{MSE} $\downarrow$ \\ 
\hline
10\% & \multicolumn{1}{c|}{151} & \multicolumn{1}{c|}{ \textcolor{falured}{\xmark}}& 14.20 & 0.038 \\
25\% & \multicolumn{1}{c|}{378} &\multicolumn{1}{c|}{ \textcolor{falured}{\xmark}} & 14.98 & 0.032 \\
50\% & \multicolumn{1}{c|}{757} & \multicolumn{1}{c|}{\textcolor{falured}{\xmark}} & 15.42 & 0.028 \\
100\% & \multicolumn{1}{c|}{1515} &\multicolumn{1}{c|}{\textcolor{falured}{\xmark}}  & 16.00 & 0.025 \\
100\% & \multicolumn{1}{c|}{1515} & \multicolumn{1}{c|}{\textcolor{teal}{\checkmark}} &  \cellcolor{tabfirst}\textbf{19.09} & \cellcolor{tabfirst}
\textbf{0.012} \\[-0.1em]
\bottomrule
\end{tabular}
\end{minipage}
% }
% \caption{\textbf{Effect of percent of pretraining scenes on voxel-grid reconstruction} showing our pretraining on larger datasets helps to learn better representations.}
% \label{tab:scaling_law_reconstruction}
% \end{table}

% \vspace{1mm}
\caption{\textbf{Effect of percent of pretraining scenes} on downstream 3D object detection (left) and voxel-grid reconstruction (right) showing more unlabeled data helps learn better representation for downstream 3D tasks as well as better reconstruction quality of unmasked patches in the input radiance and density grid} 
\label{tab:pretraining_scene_analysis}
% \vspace{-8mm}
\end{table}

% \vspace{-0.1mm}
\captionsetup[table]{skip=2pt}
\begin{table}[ht!]
% \vspace{-2mm}
% \renewcommand{\arraystretch}{0.8}
% \begin{table}[b!]
% \centering
% \scriptsize

\begin{minipage}{.56\linewidth}
\scriptsize
\setlength{\tabcolsep}{0.6em}
\centering

% \resizebox{0.45\textwidth}{!}{%
\begin{tabular}{ccccc}
\toprule
& \multicolumn{2}{c}{Front3D~\cite{hu2023nerf}} & \multicolumn{2}{c}{ScanNet~\cite{dai2017scannet}}\\
\cmidrule(lr){2-5}
NeRF-MAE& \textbf{Recall@50}$\uparrow$ & \textbf{AP@25}$\uparrow$ & \textbf{Recall@50}$\uparrow$ & \textbf{AP@25}$\uparrow$ \\

\midrule
 \multicolumn{1}{c|}{w/o Skip}  & 0.73 &  \multicolumn{1}{c|}{0.83}  & 0.39 & 0.54 \\
 \multicolumn{1}{c|}{w/ Skip}  & \cellcolor{tabfirst}\textbf{0.74} &  \multicolumn{1}{c|}{\cellcolor{tabfirst}\textbf{0.85}}   & \cellcolor{tabfirst}\textbf{0.40} & \cellcolor{tabfirst}\textbf{0.57} \\
\bottomrule
\end{tabular}
% }

% \caption{\textbf{Effect of skip connections in the decoder} during pretraining showing NeRF-MAE with skip connection helps learn better representations for downstream tasks}
% \label{tab:skip_ablation}
% \end{table}

\end{minipage}
\hspace{1mm}
\begin{minipage}{.46\linewidth}
\scriptsize
\setlength{\tabcolsep}{0.8em}
\centering

% \resizebox{0.45\textwidth}{!}{%
\begin{tabular}{ccc}
\toprule
\multicolumn{3}{c}{\textit{Masking Ratio Ablation}}\\
\midrule \\
[-1.5em]
Method & \textbf{AP@25}$\uparrow$ & \textbf{AP@50}$\uparrow$ \\
\hline
\multicolumn{1}{c|}{65\%}  & 0.805 & 0.519   \\
\multicolumn{1}{c|}{75\%} & \textbf{0.839} & \textbf{0.546} \\
\multicolumn{1}{c|}{85\%} & 0.814 & 0.512 \\
\bottomrule
\end{tabular}
% }

% \caption{\textbf{Effect of skip connections in the decoder} during pretraining showing NeRF-MAE with skip connection helps learn better representations for downstream tasks}
% \label{tab:skip_ablation}
% \end{table}

\end{minipage}
% \vspace{1mm}
\caption{\textbf{Effect of skip connections in the decoder (Left) and masking ratio ablation (Right):} shows NeRF-MAE with skip connection helps learn better representations for downstream tasks as well as 75\% masking ratio achieves the best downstream performance. }
\vspace{-8mm}
\label{tab:skip_depth_ablation}
\end{table}

\section{Experiment Analysis}
\label{sec:add_experiments}

In this section, we present additional experimental analysis. First, we present in detail an additional discussion on the downstream 3D object detection, semantic voxel-labeling, and voxel super-resolution tasks in Section~\ref{subsec:further_comp_downstream}. We present the additional analysis of scaling laws and ablation studies in Section~\ref{subsec:addition_ablate} specifically highlighting the architectural ablation of adding skip connections to the decoder and discussing masking ratio ablation. We present additional insights with comparison to 3D pretraining baselines in Section~\ref{subsec:addition_pretraining_analysis} and details about per-scene NeRF optimization to prepare our pretraining dataset in Section~\ref{subsec:pertaining_per_scene_optimization}

\subsection{Comparison on Downstream 3D Tasks} 
\label{subsec:further_comp_downstream}
We analyze all of the improvement numbers mentioned in Table~\ref{tab:downstream_3d_obb} and Table~\ref{tab:comparison_downsteam_voxel_superresolution} by highlighting them in Figure~\ref{fig:comparison_chart2}. One can clearly see our pretraining consistently improves all metrics across different datasets, especially on ScanNet which is omitted from our pretraining and used as a hold-out dataset for our cross-dataset transfer experiments. The results also clearly show that downstream 3D task numbers for 3D OBB prediction improve with adding unlabelled data from diverse sources, in this case, adding unlabelled HM3D scenes to our pretraining improves ScanNet numbers showing the efficacy of our pretraining strategy. Note that we only utilized posed 2D data for pertaining, and no 3D information was used in our pretraining pipeline. 

Next, we present results for the task of voxel grid super-resolution on HM3D dataset hold-out scenes never seen by the model during pretraining or transfer learning. We summarize these results in Table~\ref{tab:comparison_downsteam_hm3d}. 
\begin{figure}[t]
\includegraphics[width=1.0\textwidth]{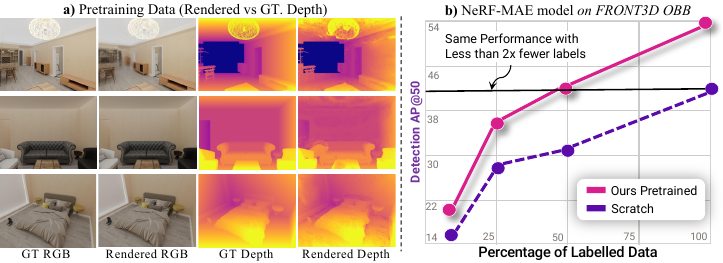}
\caption{\textbf{Qualitative Rendered Depth (a) and Quantitative Detection Performance with a varying number of labeled scenes (b):} We show the qualitative comparison of ground truth vs rendered depth map from NeRFs used for our pretraining scenes showing faithful reconstruction of geometry for our input data. We further show in the chart on the right~(b), that our pretraining requires much less data to achieve the same performance as the strong baseline.}
\label{fig:depth_percent_of_labelled}
\end{figure}

One can clearly observe that our technique achieves a higher Recall@50 and AP@50 compared to strong baseline, specifically showing improvement of 9\% in Recall@50 and 7\% in AP@50, further confirming the efficacy of our pretraining strategy on a challenging dataset and 3D task. The results in Table~\ref{tab:prior_approaches_comparison} (Left) further confirm the finding of Table~\ref{tab:comparison_downsteam_voxel_superresolution} (Left), achieving consistently higher 3D PSNR and lower MSE metrics for hold-out test scenes on HM3D for the challenging voxel grid super-resolution task.  

We highlight additional qualitative and qualitative comparison results in Figure~\ref{fig:depth_percent_of_labelled}. This figure shows the rendered depth maps against ground truth depth which highlights the input used to pretrain our representation provides good geometric cues to the network. We further show the impact of our pretraining on the amount of labeled data required during finetuning. One can clearly see from Figure~\ref{fig:depth_percent_of_labelled}~(Right) that our pretraining requires twice as fewer scenes for finetuning compared to the state-of-the-art baseline that starts from scratch to achieve the same performance, especially achieving an AP50 of 0.42 with only 50\% of the labeled data for finetuning~(i.e. 61 scenes) whereas the strong baseline uses 100\% of the labeled data~(i.e. 122 scenes) to achieve a similar performance with 2x as much data as required by our pipeline.

\begin{figure}[t!]
\centering
\resizebox{1.0\linewidth}{!}{%
\includegraphics{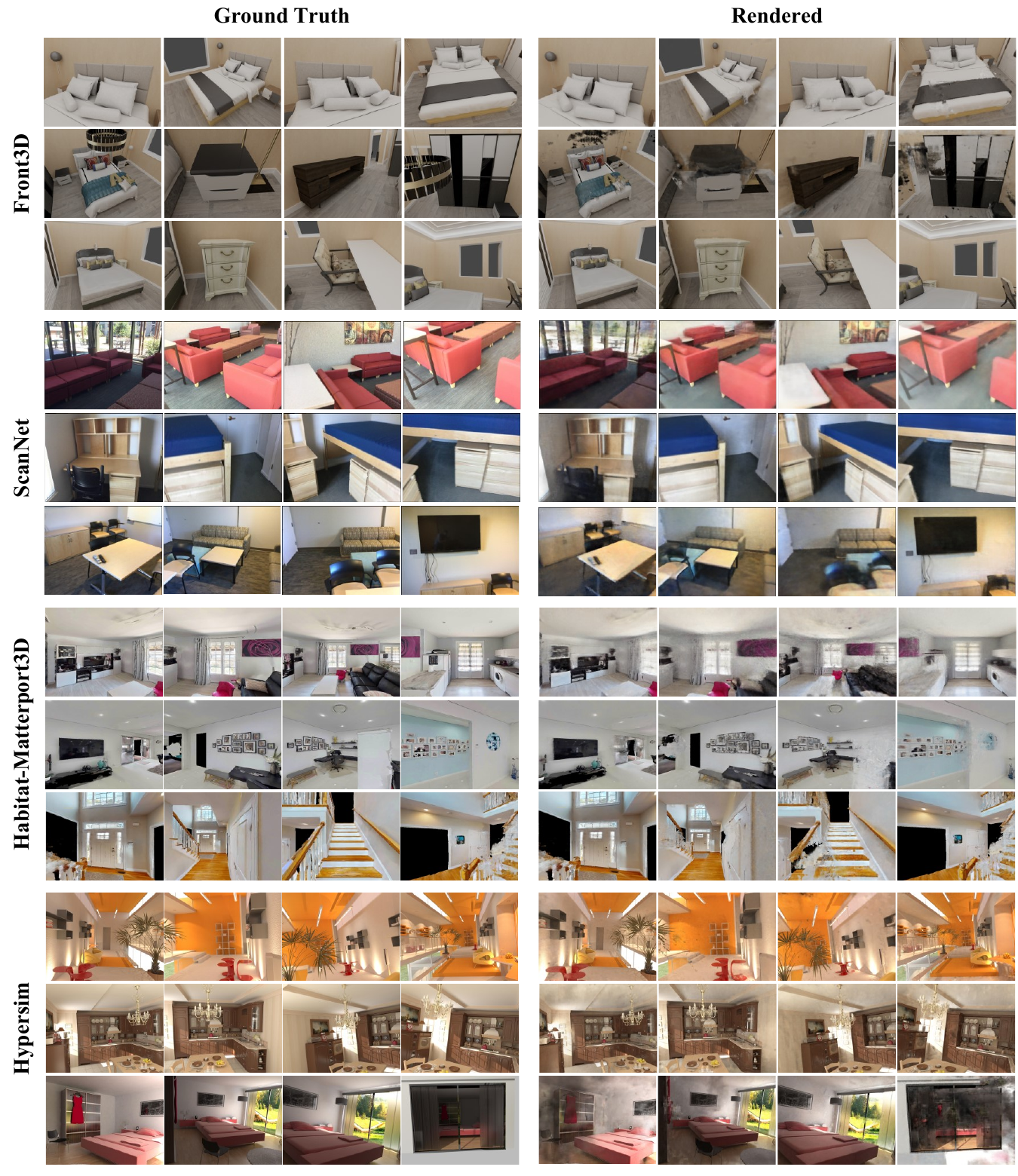}}
\captionof{figure}{
\textbf{Multi-view Rendered vs Ground-truth posed RGB} images from four distinct datasets used in this paper i.e. Front3D~\cite{fu20213d}, ScanNet~\cite{dai2017scannet}, Habitat-Matterport3D~\cite{ramakrishnan2021hm3d} and Hypersim~\cite{roberts2021hypersim}. As mentioned in the main paper, we use Instant-NGP to train neural radiance fields for Front3D, Hypersim and HM3D, whereas Scannet scenes are trained using dense depth prior nerf~\cite{roessle2022dense}, following~\cite{hu2023nerf}
}
\label{fig:nerf_data}
\end{figure}

\subsection{Additional Ablation Analysis} 
\label{subsec:addition_ablate}
We report complete metrics for our scaling laws experiments shown in Figure~\ref{fig:quantitative_results_scaling_laws} of the main paper. We report these metrics in Table~\ref{tab:pretraining_scene_analysis}. Both results further confirm our findings in the paper that our results consistently improve on reconstruction task as well as during transfer learning by utilizing more unlabelled data during pretraining. We additionally include ablations to our NeRF-MAE pretraining architecture in Table~\ref{tab:skip_depth_ablation}. We show the effect of adding skip connections to the lightweight decoders in Table~\ref{tab:skip_depth_ablation}~(Left). One can clearly see that adding skip connections during reconstruction increases the accuracy of downstream tasks, specifically increasing the AP25 by 2\% and 3\% on FRONT3D and ScanNet datasets respectively. We also analyze the effect of masking ratio on downstream task performance in Table~\ref{tab:skip_depth_ablation}~(Right). This table clearly shows our design choice of using 75\% as the masking ratio for input radiance and density grid. This significantly increases the accuracy of 3D object detection downstream task.

\subsection{Comparison to 3D pretraining baselines} 
\label{subsec:addition_pretraining_analysis}

We present a detailed comparison to 3D pretraining baselines~\cite{pang2022masked, zhang2022point, zhang_depth_contrast, hou2021exploring} in Table~\ref{tab:prior_approaches_comparison} in our main paper. Here, we share more insights on these results. Despite given NeRF rendered depth maps~(from the same NeRF model, NeRF-MAE uses to sample its radiance and density grid from) and ground truth poses, the baselines~\cite{zhang_depth_contrast, hou2021exploring} struggle to improve performance in an extreme data-scarcity setting~(i.e.~on cross-domain generalization where 90 scenes are available for finetuning), whereas, our approach, NeRF-MAE excels in these challenging scenarios. We conclude that this is due to the local training structure of DepthContrast and other MAE approaches, whereas NeRF-MAE pretraining captures both the local and global structure of the full-scenes with its Transformer-inspired architecture. Despite given NeRF's radiance and density grid as an input to point-based pretraining baselines~\cite{pang2022masked, zhang2022point}, we see degraded performance. We conclude that these approaches use specialized modules like KNN used in Point-MAE~\cite{pang2022masked}, DGCNN~\cite{dgcnn} used in Point-BERT~\cite{yu2022point} which are specialized for modeling surface level information and require extra parameter tunings such as group size and number of centers. Even though PointM2AE~\cite{zhang2022point} improved performance over PointMAE~\cite{pang2022masked} using a hierarchical training, we show that they still struggle for this challenging downstream task where only a few labeled scenes are available for finetuning. On the contrary, our approach utilized standard Transformer modules and shows very effective downstream performance with an opacity and radiance-aware masked reconstruction objective.

\subsection{Per-scene NeRF optimization Analysis} 
\label{subsec:pertaining_per_scene_optimization}

Our per-scene NeRF optimization takes 8 mins on 1 GPU for a total of just over 2 days on 8 GPUs for the entire pre-training dataset. A by-product of this process that we would like to highlight is the compression of the original data~(see Table~\ref{tab:compression}). Our results indicate significant memory savings, and the resulting compressed representation still conveys all the necessary information for downstream tasks.

\begin{table}[t!]
\centering
\setlength{\tabcolsep}{2.5pt}
\scriptsize
\begin{tabular}{ccccc}
\toprule

& \multicolumn{2}{c}{HM3D~\cite{ramakrishnan2021hm3d}~(NeRF-MAE)} & \multicolumn{2}{c}{ScanNet~\cite{dai2017scannet} test (NeRF-MAE)}\\
\cmidrule(lr){1-2}
\cmidrule(lr){2-3} \cmidrule(lr){4-5}
Method & \textbf{Memory}$\downarrow$& \textbf{Memory(Rel)} $\downarrow$&  \textbf{Memory}$\downarrow$& \textbf{Memory(Rel)} $\downarrow$ \\ 
\midrule

\multicolumn{1}{c|}{Original~(RGB)}  & 256GB & \multicolumn{1}{c|}{$\pm$ 0.00\%}  & 30GB &$\pm$ 0.00\% \\
\multicolumn{1}{c|}{Compressed (Weights)}  &\textbf{46GB} &\multicolumn{1}{c|}{\textbf{$-$ 82.03\%}} 
  &\textbf{1.8GB} & \textbf{\textbf{$-$ 94.00\%}}  \\
\bottomrule
\end{tabular}
\caption{\textbf{Relative Pretraining Dataset Comparison:} showing significant memory saving during data preprocessing for NeRF-MAE.}
\label{tab:compression}
\end{table}

\section{Pretraining Datasets}
\label{sec:pretraining_datasets}

We show our pretraining datasets in detail in Figure~\ref{fig:pretraining_datasets} and Figure~\ref{fig:nerf_data}. First, we show the data processing flow for each of the Front3D~\cite{fu20213d}, Scannet~\cite{dai2017scannet}, HM3D~\cite{ramakrishnan2021hm3d} and Hypersim~\cite{roberts2021hypersim} dataset in Figure~\ref{fig:pretraining_datasets} starting with multi-view images to get a NeRF trained for each scene and finally getting an explicit radiance and density grid for all scenes. We train each scene for 100,000 steps to ensure high-quality NeRF are obtained as input to our pretraining stage as further confirmed by our experiments in Figure~\ref{fig:quantitative_results_scaling_laws} in the main paper, which quantitatively shows that better NeRF quality results in better pretraining and leads to improved downstream 3D task performance. We further show rendered RGB vs ground-truth RGB for selected scenes from all four datasets in Figure~\ref{fig:nerf_data}. The figure confirms that our NeRF quality for validation images is faithful to the ground-truth images. Secondly, it shows that our pretraining dataset for training good 3D representations from posed 2D data is very diverse in realism, lightning, shadows, number of objects in the scene as well as the setting of each scene i.e. ranging from bedroom scenes to office and living room scenes. This makes our 3D representation learning more robust to the varying input data and helps it improve performance for data outside the training data-distribution, as further confirmed in Section~\ref{sec:experiments}. 

\section{Compute Resources}
\label{sec:compute_resources}

Our model was trained on 7 Nvidia A100 GPUs for 1200 epochs for 2 days and 17 hours. The cost for a pretraining run is around 19 GPU days with a GPU capacity of 82GB and an average utilization of 60\%. To pretrain our representation, each NeRF was trained for 8 minutes on a single GPU for 100,000 steps. The total training time for training 3500 NeRF was 58 hours and finished in just over 2 days on 8 Nvidia A100 GPUs. Note that our pretraining strategy is agnostic to the NeRF technique used and can be replaced with any of the faster NeRF techniques developed. We show that our pretraining strategy equally applies to both instant-NGP~\cite{muller2022instant} trained NeRF and dense depth prior NeRF~\cite{roessle2022dense} and doesn't require any modification to our network architecture or training strategy.
\section{Limitations and Future Work}
\label{sec:future_work}

NeRF-MAE enables large-scale self-supervised pertaining of 3D representations using NeRF's unique radiance and density grid as an input hence employing standard 3D Transformer blocks to significantly improve various downstream 3D tasks. Although powerful, future work is still required to improve the efficiency of training this representation to allow for seamless and efficient input of a large variety of multi-view data from diverse data sources. This could be achieved using fast linear attention blocks~\cite{wang2020linformer} or employing a 3D version of state space models~\cite{fu2023hungry}. Another avenue for future work is communication between neural rendering and masking. Concurrent works have combined neural rendering with 3D representation learning~\cite{huang2023ponder, zhu2023ponderv2} albeit in a non-masking manner. Other works have used correspondences to mine multi-view image pairs to train MAE~\cite{he2021masked} and designed the task of cross-view completion~\cite{croco} for unsupervised 3D representation learning from 2D images. Our work NeRF-MAE bridges the 2D and 3D domains by using Neural Radiance Fields and future work could look into backpropagating the network weights through both our reconstructed grid as well as neural rendered images to further improve performance. Another promising avenue for future work is to explore various downstream 3D tasks that could benefit from dense NeRF-MAE pertaining such as 3D scene reconstruction~\cite{yan2022implicit, Peng2020ECCV}. 
\end{document}